\documentclass{article}

\usepackage[nonatbib, preprint]{neurips_2024}

\usepackage[utf8]{inputenc} 
\usepackage[T1]{fontenc}    
\usepackage{hyperref}       
\usepackage{url}            
\usepackage{booktabs}       
\usepackage{amsfonts}       
\usepackage{nicefrac}       
\usepackage{microtype}      
\usepackage{xcolor}         
\usepackage{graphicx}
\usepackage{caption}
\usepackage{subcaption}
\usepackage{multirow}
\usepackage{xcolor,colortbl}
\usepackage{pifont}
\usepackage{amsmath}

\definecolor{maroon}{cmyk}{0.39, 0, 0.39, 0.07}

\title{TableVQA-Bench: A Visual Question Answering Benchmark on Multiple Table Domains}

\author{%
  Yoonsik Kim\thanks{Corresponding author} \\
  NAVER Cloud AI \\ 
  Seongnam-si, Gyeonggi-do, Korea \\
  \texttt{yoonsik.kim90@navercorp.com} \\
  \And
  Moonbin Yim \\
  NAVER Cloud AI \\ 
  Seongnam-si, Gyeonggi-do, Korea \\
  \texttt{moonbin.yim@navercorp.com} \\
  \AND
  Ka Yeon Song\\
  NAVER Cloud AI \\ 
  Seongnam-si, Gyeonggi-do, Korea \\
  \texttt{kayeon.song@navercorp.com} \\  
}

\begin{document}

\maketitle

\begin{abstract} 
In this paper, we establish a benchmark for table visual question answering, referred to as the TableVQA-Bench, derived from pre-existing table question-answering (QA) and table structure recognition datasets. 
It is important to note that existing datasets have not incorporated images or QA pairs, which are two crucial components of TableVQA.
As such, the primary objective of this paper is to obtain these necessary components.
Specifically, images are sourced either through the application of a \textit{stylesheet} or by employing the proposed table rendering system. 
QA pairs are generated by exploiting the large language model (LLM) where the input is a text-formatted table. 
Ultimately, the completed TableVQA-Bench comprises 1,500 QA pairs. 
We comprehensively compare the performance of various multi-modal large language models (MLLMs) on TableVQA-Bench. GPT-4V achieves the highest accuracy among commercial and open-sourced MLLMs from our experiments. 
Moreover, we discover that the number of vision queries plays a significant role in TableVQA performance. To further analyze the capabilities of MLLMs in comparison to their LLM backbones, we investigate by presenting image-formatted tables to MLLMs and text-formatted tables to LLMs, respectively. 
Our findings suggest that processing visual inputs is more challenging than text inputs, as evidenced by the lower performance of MLLMs, despite generally requiring higher computational costs than LLMs. 
The proposed TableVQA-Bench and evaluation codes are available at \href{https://github.com/naver-ai/tablevqabench}{https://github.com/naver-ai/tablevqabench}.

\end{abstract}

\section{Introduction}
Tabular data is one of the most prevalent formats for representing structured text, playing a significant role in the efficient delivery of text-based information.
A large proportion of these tables can be found in image form, created from text sources, such as HTML and markdown formats. 
Therefore, understanding visual tabular data can be deemed a crucial endeavor within the realm of the visual documentation domain.
In light of recent advancements in multi-modal large language models (MLLMs)~\cite{llava,blip2,zhu2023minigpt,cream,lin2023sphinx}, it is now possible to harbor this capability within a single model. However, despite its significance, the evaluation of visual table data has been less vigorous due to the absence of evaluation datasets.

Meanwhile, in natural language processing (NLP), textual table question answering (TableQA) datasets have been widely proposed. 
For instance, Panupong \emph{et al.} provide WikiTableQuestion (WTQ)~\cite{wtq} that is a question-answering task based on a text-based table. Chen \emph{et al.} also release TabFact~\cite{tabfact} dataset determining whether a statement is entailed or refuted with a given table. Unfortunately, these datasets do not provide table images, making it challenging to apply them directly to table visual question answering.

\begin{figure}[t]
    \centering   
    \includegraphics[width=0.95\textwidth]{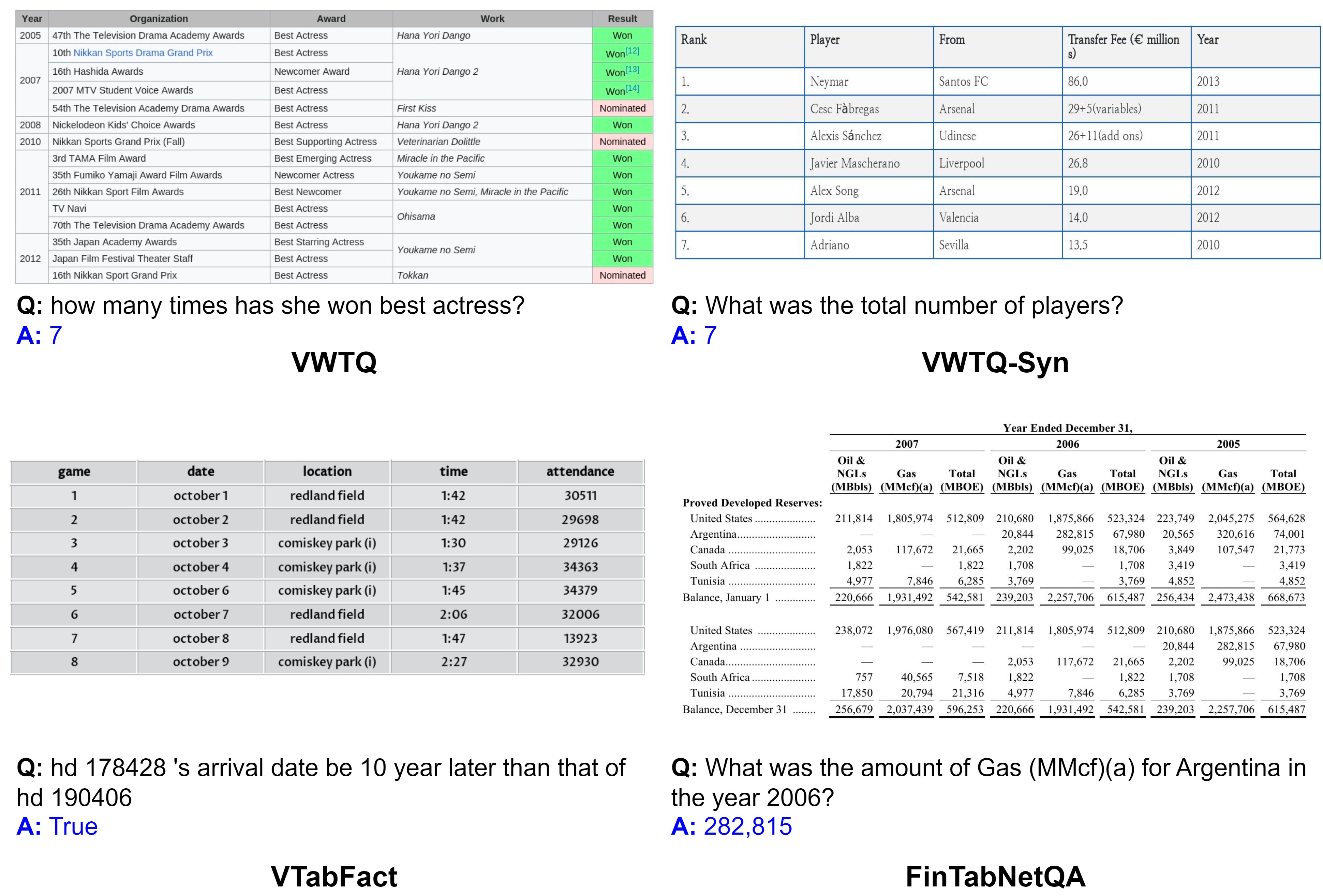}    
    \caption{Samples of the proposed TableVQA-Bench. TableVQA-Bench incorporates four domains of table datasets: VWTQ, VWTQ-Syn, VTabFact, and FinTabNetQA. The images of VWTQ-Syn and VTabFact are generated by our rendering system.}
    \label{fig:samples}  
\end{figure}

In this paper, we construct a new TableVQA-Bench dataset as shown in Fig.~\ref{fig:samples} by leveraging existing TableQA and table structure recognition (TSR) datasets. 
As for the TableQA dataset, real table images are sourced by attaching a \textit{stylesheet} of original source (Wikipedia) into HTML that contains both the content and style of the table.  
Acquired images can be contaminated, given that Wikipedia is often utilized as a primary source for constructing the web-crawled base for pre-training data, as suggested by Pix2Struct~\cite{lee2023pix2struct}. To circumvent this issue, the proposed table rendering system is also utilized to obtain synthetic table images. 
As for TSR dataset, QA pairs are required for constructing TableVQA. 
To generate QA pairs, we propose to exploit GPT-4~\cite{gpt4} by feeding the text-formatted table as an input.

Through comparisons among MLLMs on TableVQA-Bench, we found that GPT-4V~\cite{gpt4v} outperforms other methods including commercial and open-sourced models across all table domains. We also observed that preserving the original information of visual features can be a crucial factor for TableVQA. For example, GPT-4V and CogVLM achieved enhanced performance when the resolution of the input image was higher.    
\begin{figure}[t]
\centering
    \begin{subfigure}[b]{1.0\textwidth}
        \centering
        \includegraphics[width=1.0\textwidth]{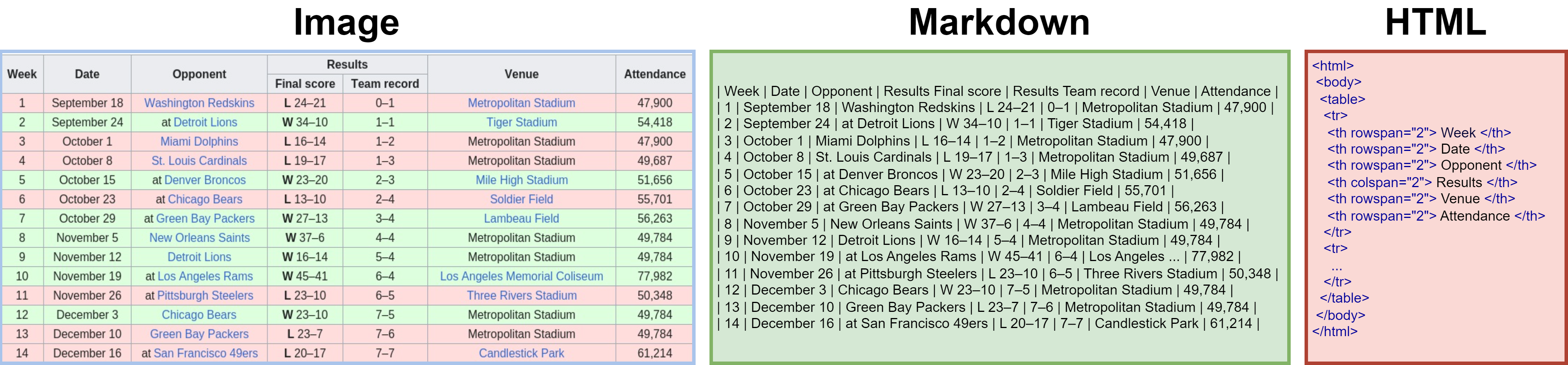}     
        \caption{Examples of Table Formats.}
        \label{fig:input_format_visualization}
    \end{subfigure} \vspace{0.5em}
    \begin{subfigure}[b]{0.8\textwidth}
        \centering
        \includegraphics[width=1.0\textwidth]{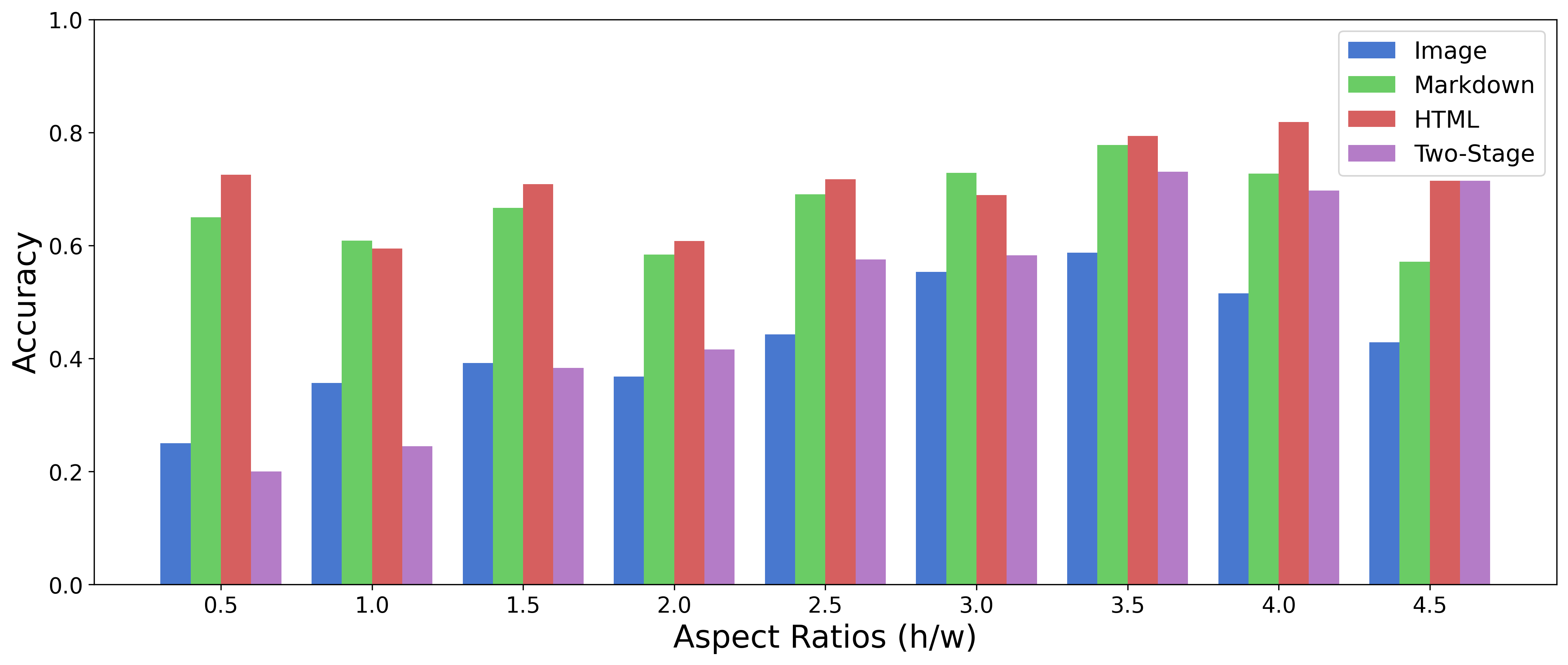}      
        \caption{Performance based on Input Formats.}
        \label{fig:input_format_performance}
    \end{subfigure}
\caption{
We present visualized examples with various formats having the same content.  
The evaluations of GPT-4 families~\cite{gpt4,gpt4v} were conducted on VWTQ, which consists of 750 samples. 
The accuracy of a vision-formatted table gets lower performance than the accuracy of a text-formatted table and severely depends on the aspect ratio of the input image.
Since HTML format effectively represents multi-row and multi-column configurations, it can achieve better performance than markdown format. 
} \label{fig:input_format} 
\end{figure}
To provide a better analysis of the model's capability, we conduct a comprehensive investigation of table formats and their performance. As illustrated in Fig.~\ref{fig:input_format}, text-formatted tables, including HTML and markdown, tended to outperform their vision-formatted counterparts. 
Furthermore, to enhance the analysis, a two-stage approach is explored, which initially involves extracting content from images for HTML representation and subsequently applying it to the TableQA task.

\section{Related Works}
The significance of benchmarks for assessing the performance of MLLMs has grown as MLLMs advance rapidly. MMBench~\cite{MMBench} evaluates perception and reasoning across approximately 3,000 questions in 20 different ability dimensions, including the `image-text understanding' dimension. SEED-Bench~\cite{SEED-Bench} is categorized into 12 evaluation dimensions with about 19,000 questions covering scenes, detection, OCR, and various other types. SEED-Bench-2~\cite{SEED-Bench-2} increases the number of questions to 24K to its predecessor, and the complexity of questions has been heightened to represent multi-modal content on both input and output sides. MathVista~\cite{MathVista} is a mathematically specialized evaluation set, consisting of 6,141 subjective and objective questions. This dataset encompasses questions related to seven types of mathematical reasoning and covers five primary tasks, incorporating a small portion in tabular format. Recently, chart question-answering benchmarks~\cite{chartqa,xu2023chartbench,liu2023mmc} have been introduced, examining specific domains of tasks. While these aforementioned datasets may partially encompass or relate to TableVQA, they do not primarily focus on TableVQA. Therefore, a dataset meticulously designed for the thorough investigation of TableVQA is indispensable and our TableVQA-Bench dutifully fulfills this requirement. Furthermore, we believe that the extensive investigation provided in this paper will be helpful in interpreting the table-related performance in previous datasets. 
\begin{figure}[h]
    \centering   
    \includegraphics[width=0.95\textwidth]{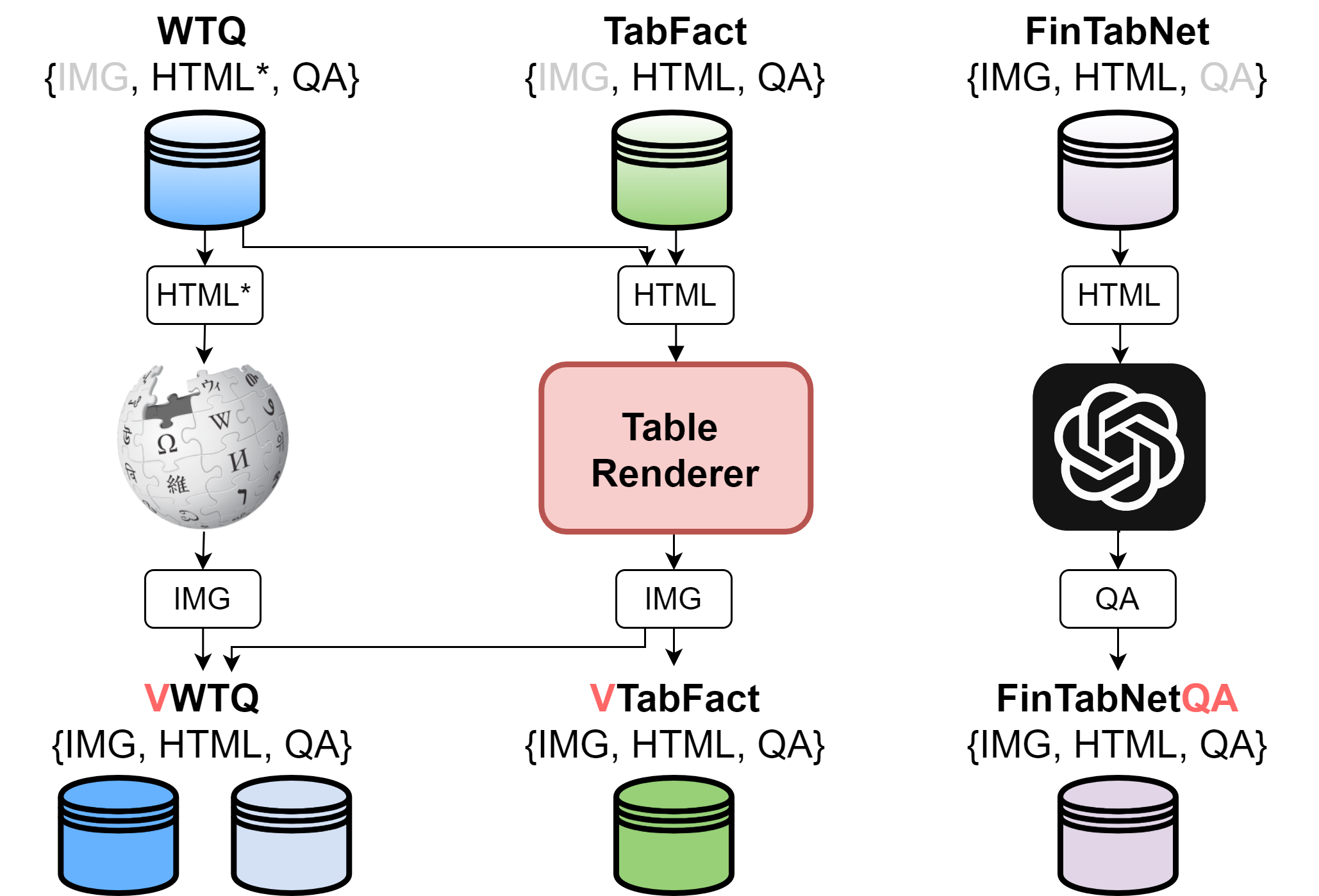}     
    \caption{Overview of constructing the proposed TableVQA-Bench. HTML* denotes that it incorporates both content and style of tables, while HTML only contains content. 
    }
    \label{fig:main} 
\end{figure}

\section{TableVQA-Bench}

As illustrated in Fig.~\ref{fig:main}, we construct the TableVQA-Bench. TableVQA-Bench encompasses VWTQ, VTabFact, and FinTabNetQA, which are extended from pre-existing databases such as WTQ~\cite{wtq}, TabFact~\cite{tabfact}, and FinTabNet~\cite{fintabnet} correspondingly. 
The components of TableVQA consist of three parts; table image, text-representation (HTML), and QA pairs $\{$\textit{IMG}, \textit{HTML}, \textit{QA}$\}$.
To acquire images for VWTQ and VTabFact, we source images by attaching the \textit{stylesheet} of Wikipedia or by utilizing our table rendering system.
Conversely, FinTabNet is devoid of the QA pair, which is generated by employing the GPT-4. 
In the final stage of these processes, any samples with more than 50 table rows are methodically filtered out and the authors carry out a meticulous review.  

\subsection{VWTQ}
\begin{figure}[h]
\centering
    \begin{subfigure}[b]{0.48\textwidth}
        \centering
        \includegraphics[width=1.0\textwidth]{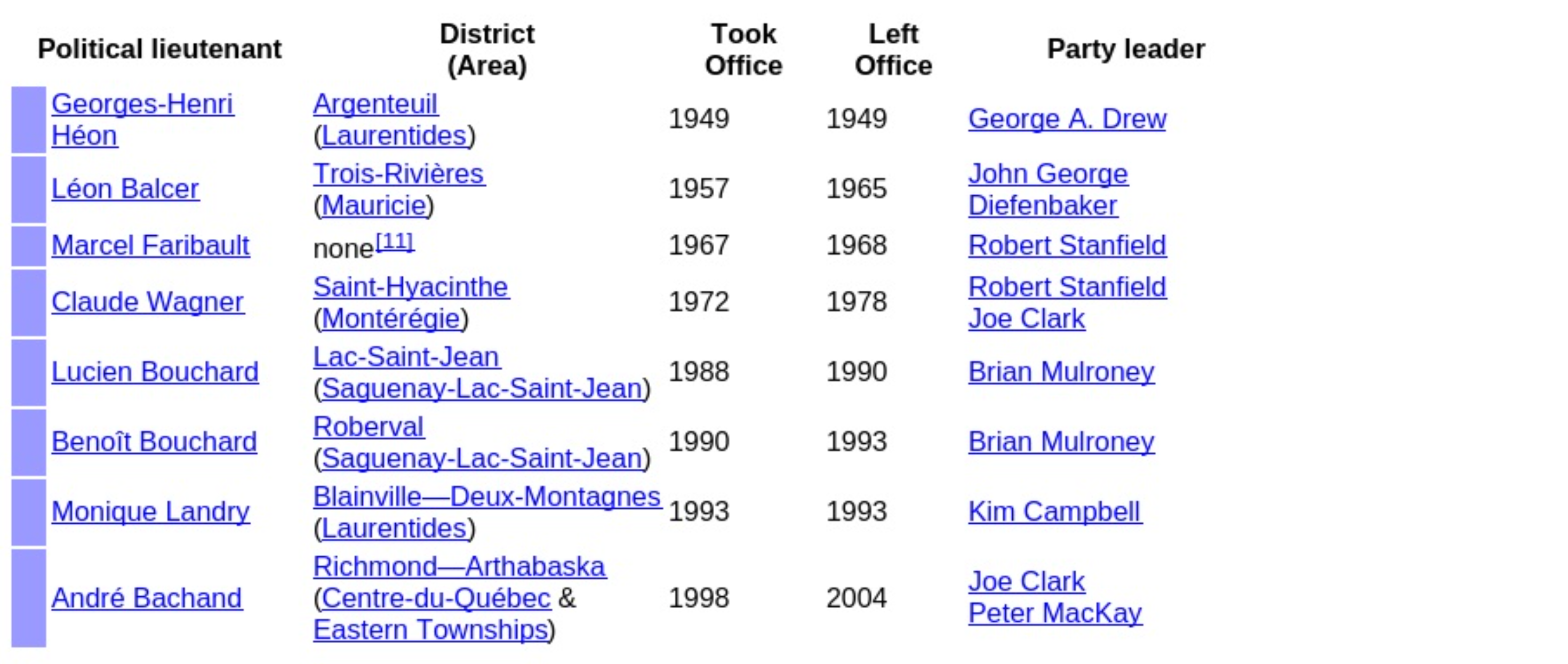}     
        \caption{Before attaching \textit{stylesheet}}
        \label{fig:before_style}
    \end{subfigure} \vspace{0.5em}
    \begin{subfigure}[b]{0.48\textwidth}
        \centering
        \includegraphics[width=1.0\textwidth]{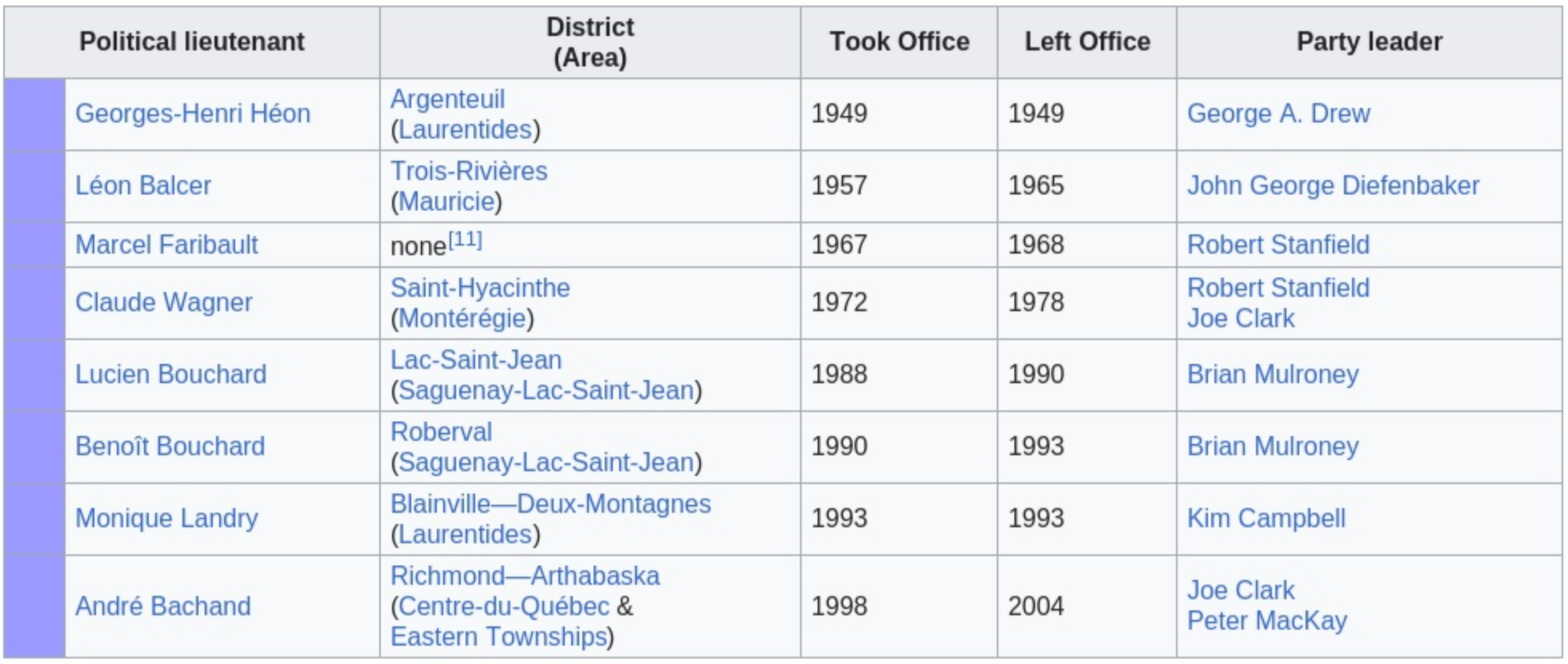}      
        \caption{After attaching \textit{stylesheet}}
        \label{fig:after_style}
    \end{subfigure}
\caption{
The captured images whether the \textit{stylesheet} is attached or not.
} \label{fig:style}
\end{figure}
VWTQ is constructed by incorporating an image collection into the WTQ~\cite{wtq} while maintaining its QA pairs and accuracy-based evaluation metric. 
As shown in Fig.~\ref{fig:before_style}, WTQ provides HTML that represents both the content and style of a table.
To reproduce the original table images from Wikipedia, we applied the \textit{stylesheet} of Wikipedia to the HTML.
Finally, we obtained the images by capturing screenshots, which are presented in Fig.~\ref{fig:after_style}.
Since images from Wikipedia can be web-crawled to gather pre-training data for MLLMs, we also generate table images using our table rendering system.
It takes HTML as input and generates tables with various styles, featuring random attributes, as detailed in Section~\ref{sec:table_renderer}.
The datasets generated from the attaching Wikipedia \textit{stylesheet} and our rendering system have been named VWTQ and VWTQ-Synthesized (VWTQ-Syn), respectively.

\subsection{VTabFact}
TabFact~\cite{tabfact} represents a verification task that verifies whether a statement derived from a table is either entailed or refuted, thus categorizing it as a variant of the TableQA task. 
In our empirical experiments, it was observed that prompts framed as ``True or False'' yielded higher efficacy compared to those framed as ``entailed or refuted''. Consequently, we replace the answer format to ``True'' or ``False'' accordingly and we employ the evaluation metric as accuracy following the TabFact. 
Given that TabFact has not provided the original HTML format of the tables, the acquisition of images is feasible only through the utilization of the proposed rendering system. It takes pseudo-HTML as an input, which is converted from the simple CSV file, and generates the images.

\subsection{FinTabNetQA}
FinTabNet~\cite{fintabnet} is a dataset for TSR task~\cite{pubtabnet,tableformer,scob} that extracts an HTML format from a given table image. Unlike WTQ and TabFact, which use Wikipedia as their data source, FinTabNet's sources are the annual reports of S\&P 500 companies, allowing it to evaluate tables from new domains. For the construction of FinTabNetQA, a generation process of QA pairs is required, and we utilized GPT-4 with HTML as an input. 
During the generation process, two issues were encountered and resolved in the following manners: 
\begin{itemize}
\item The first question is often answered in the first non-header cell.
This issue persisted even with the use of additional instructions, thus we opted to generate numerous QA pairs from a single table and conducted random sampling from QA pairs.  
\item We observed inconsistent inclusion of scale units, such as thousand, million, and billion at the answer. Particularly when the scale unit is in thousands, most generated answers often do not include the scale unit.
We rectify this issue with a meticulous human revision procedure.
\end{itemize}

In terms of the evaluation metric, we employ accuracy. It should be noted that the majority of financial tables encompass scale units, for instance, thousand, million, billion, trillion, and percentage.
For the FinTabNetQA, the accuracy measure referred to as \textit{relieved-accuracy} is employed, whereby these units are intentionally excluded during evaluation. 
To provide an illustrative example, when the ground truth is ``128 million'', predictions such as ``128 million'', ``128,000,000'' and ``128'' are all approved as accurate responses. 
This methodology is justified due to the fact that MLLMs presently fail to attain substantial performance in a strict accuracy evaluation. 
Both the \textit{strict-accuracy} and the \textit{relieved-accuracy} scripts will be made available for further research.

\begin{figure}[t]
    \centering   
    \includegraphics[width=1.0\textwidth]{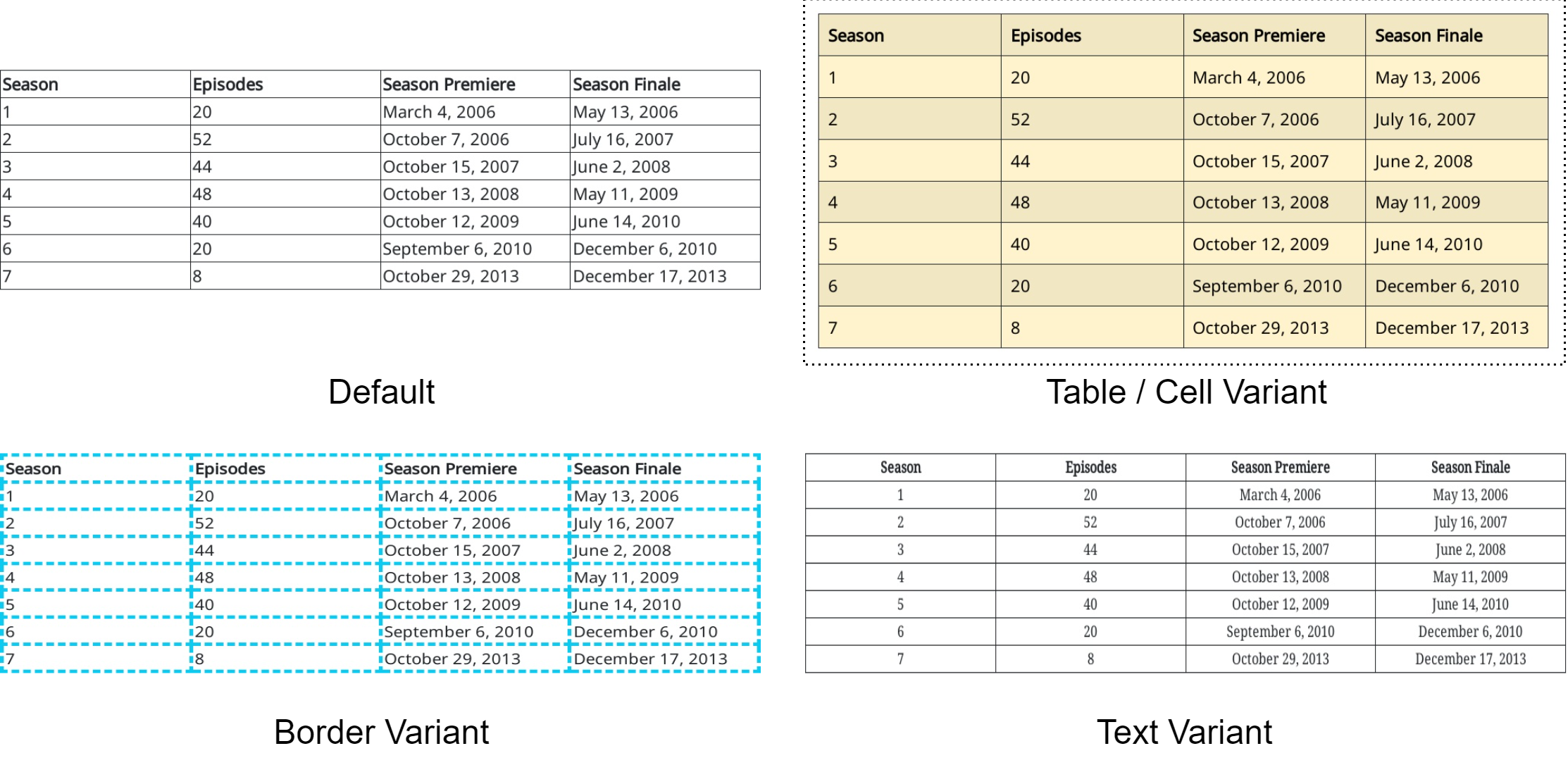}    
    \caption{The generated image according to the change of the attributes. To represent the table's margin, we denote the dashed-box as the captured table image with a white margin.}
    \label{fig:render_options}  
\end{figure}

\subsection{Table Rendering System} \label{sec:table_renderer}
Our rendering framework employs a rule-based methodology for rendering table images, engaging diverse styles applied to HTML sources. 
This framework bifurcates into two principal phases: style generation and image generation.

In the first stage, style tags are added to the original HTML to generate a styled HTML where the most of original HTML only incorporates the structure of the table. Leveraging the Bootstrap framework\footnote{https://getbootstrap.com}, the system facilitates a diverse representation of table styles encompassing elements such as cells, borders, and texts. The specific style attributes include:
\begin{itemize}
\item \textbf{Table}: background-color and margin
\item \textbf{Cell}: background-color and padding
\item \textbf{Border}: border-width, border-style, and border-color
\item \textbf{Text}: font-family, font-size, text-align, and color
\end{itemize}
where these components are randomly determined. 
Fig.~\ref{fig:render_options} presents the example when each attribute is changed from the default setting. 
The second phase, image generation, involves rendering the styled HTML within a web browser to capture a screenshot. Utilizing the Puppeteer library\footnote{https://pptr.dev}, we obtain rendered images by randomly selecting parameters such as image dimensions and JPEG quality.
To generate diverse table images, most attributes are randomly determined. However, certain attribute combinations may yield images that appear unnatural. To mitigate this, a human review process is conducted to filter out such anomalous images.

\setlength{\tabcolsep}{4pt}
\renewcommand{\arraystretch}{1.3} 
\begin{table}[t]

\begin{center}
\caption{Statistics of TableVQA-Bench.} \label{table:data_abstract}
\resizebox{0.7\linewidth}{!}{%
\begin{tabular}{lcccc} \toprule
            & Real Image & Human Generated QA & \#Image & \#QA \\ \midrule \midrule
VWTQ         & \ding{51}         & \ding{51}               &   315      & 750   \\
VWTQ-Syn     &  \ding{55}      & \ding{51}               &     150     & 250   \\
VTabFact &   \ding{55}       & \ding{51}              &    224      & 250   \\
FinTabNetQA   & \ding{51}          &  \ding{55}              &    205      & 250  \\ \midrule
Total       &    -        &   -              &    894      & 1,500 \\ \bottomrule  

\end{tabular}} 
\end{center}
\end{table}\vspace{-0.5em}

\begin{figure}[h!]
\centering
    \begin{subfigure}[b]{1.0\textwidth}
        \centering
        \includegraphics[width=0.24\textwidth]{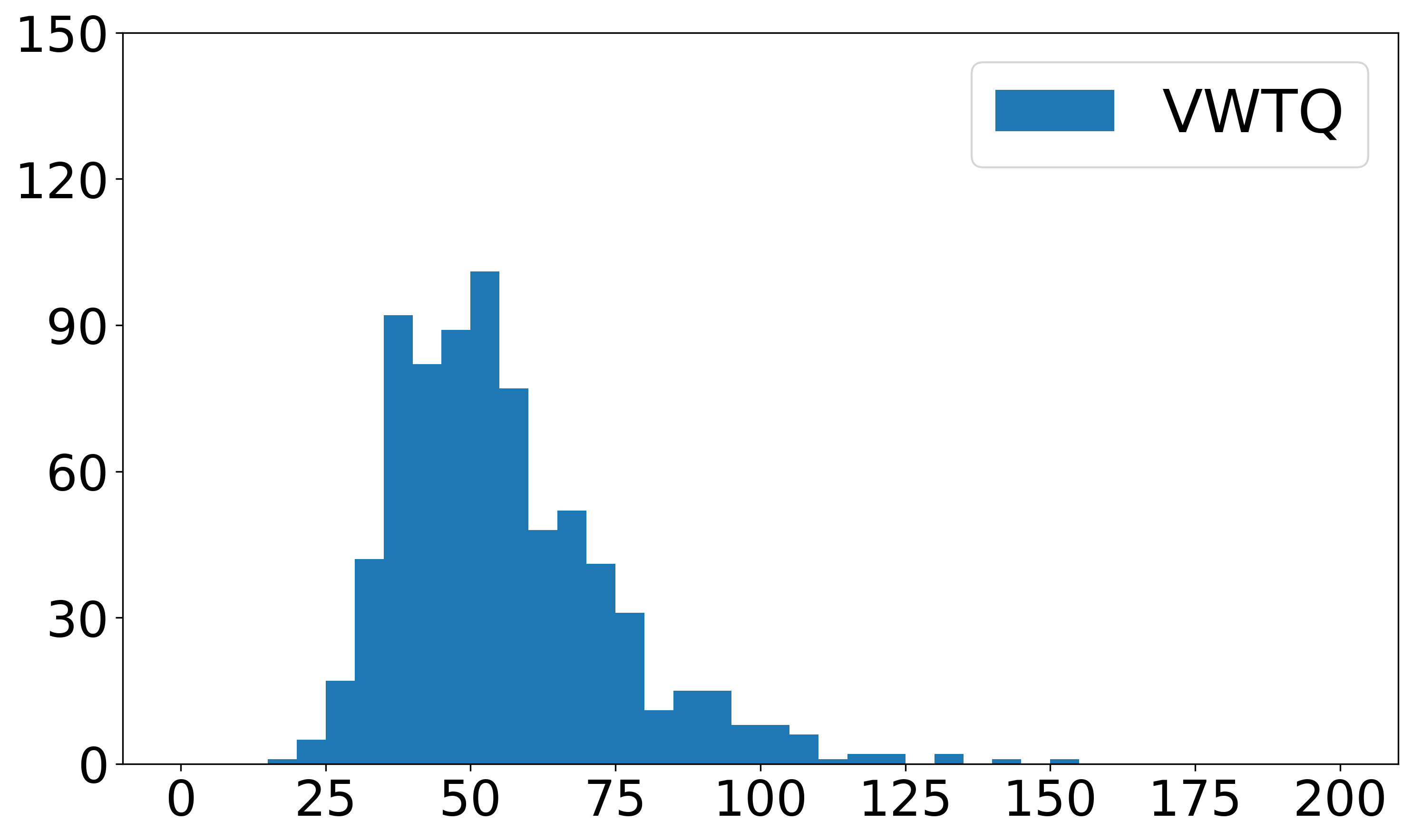}
        \includegraphics[width=0.24\textwidth]{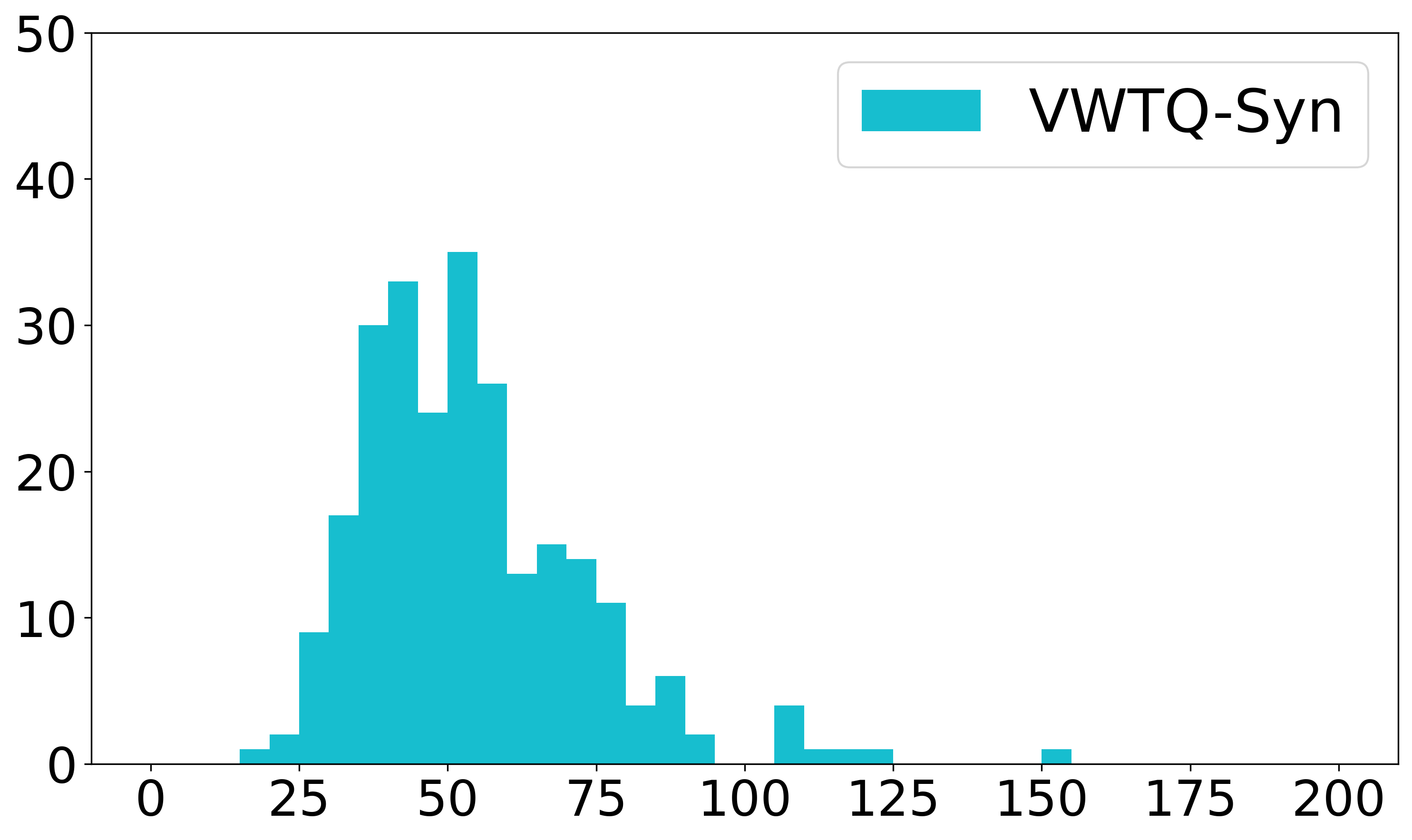} 
        \includegraphics[width=0.24\textwidth]{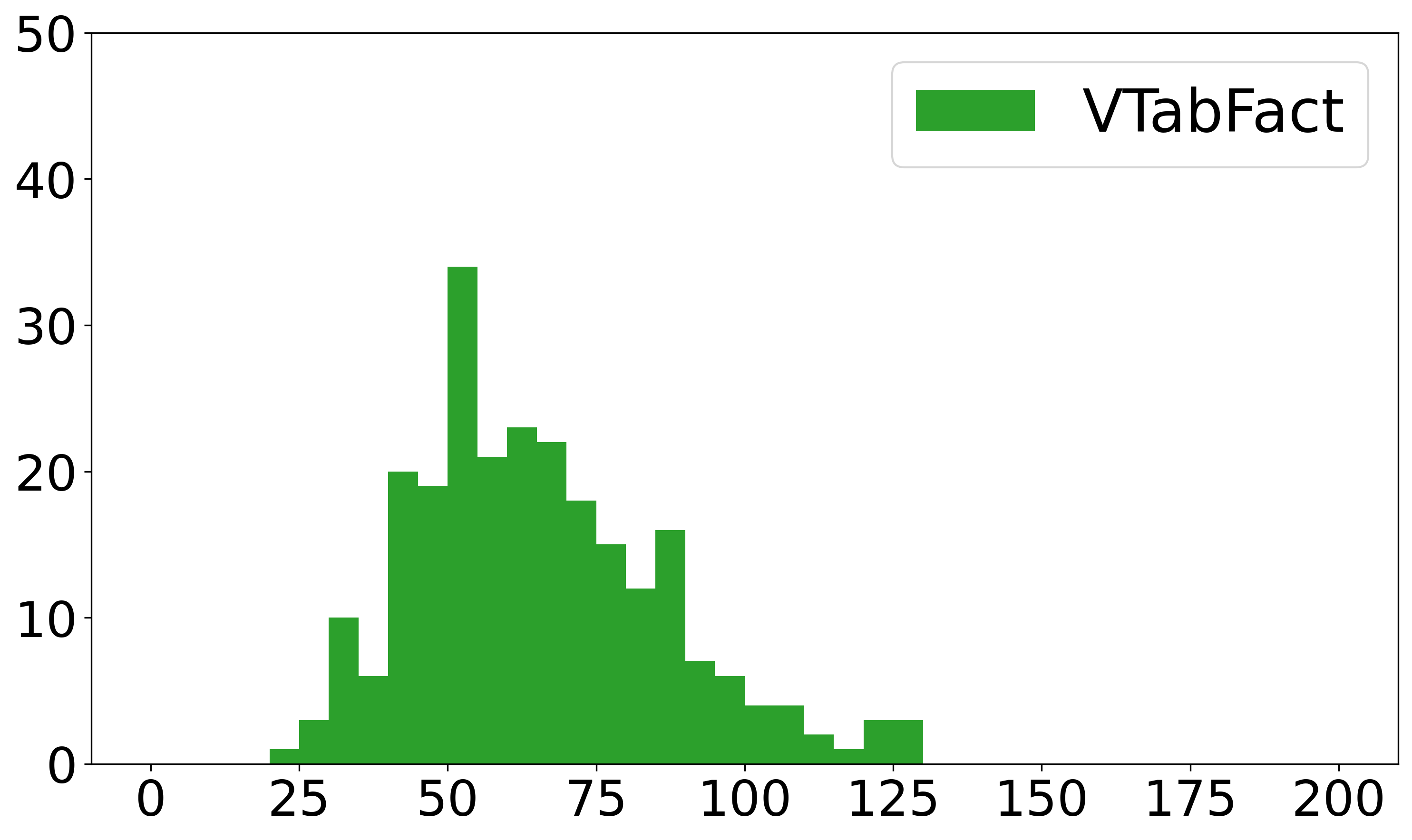} 
        \includegraphics[width=0.24\textwidth]{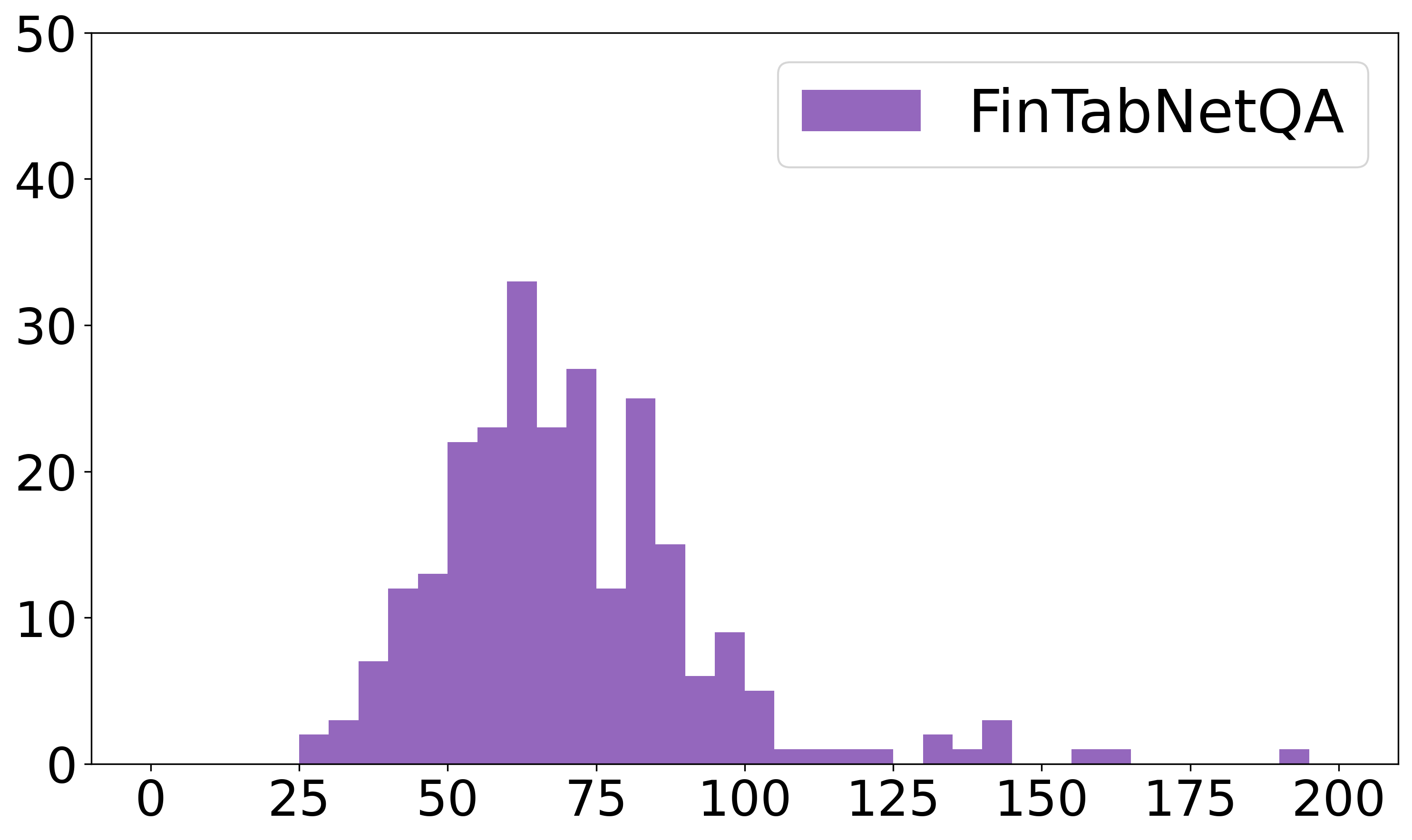} 
        \caption{Question Length}
        \label{fig:eda_question}
    \end{subfigure} 
    \begin{subfigure}[b]{1.0\textwidth}
        \centering
        \includegraphics[width=0.24\textwidth]{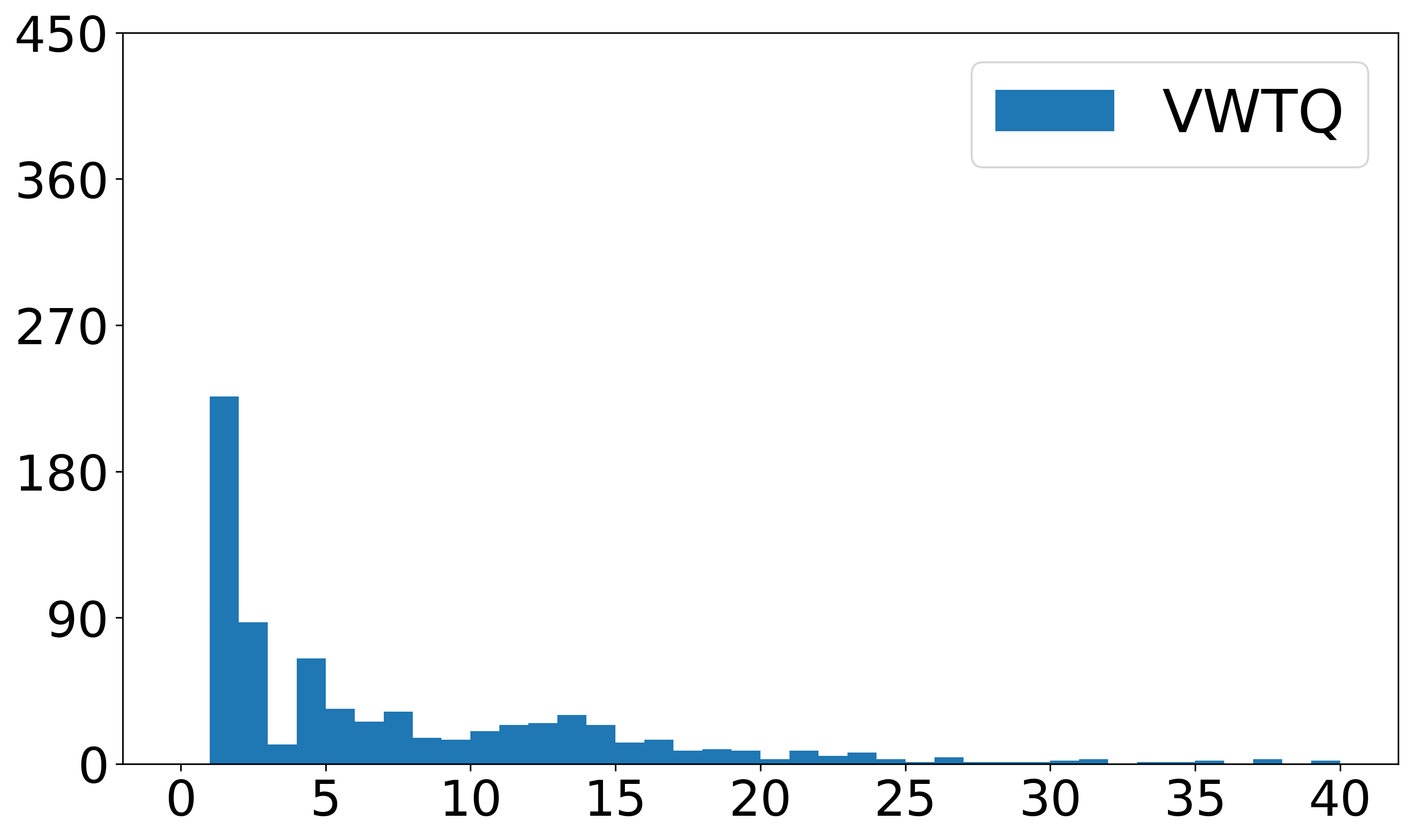}
        \includegraphics[width=0.24\textwidth]{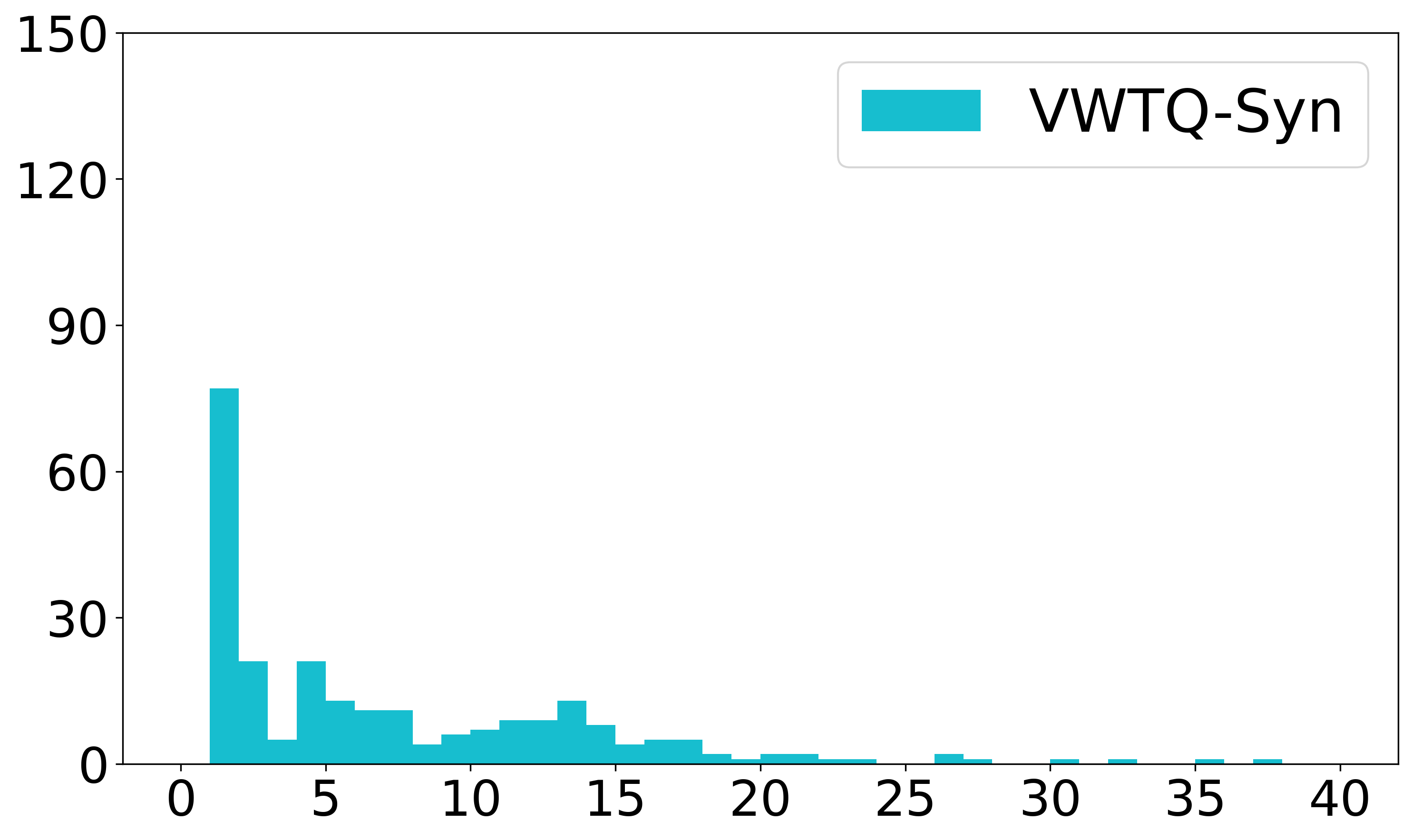} 
        \includegraphics[width=0.24\textwidth]{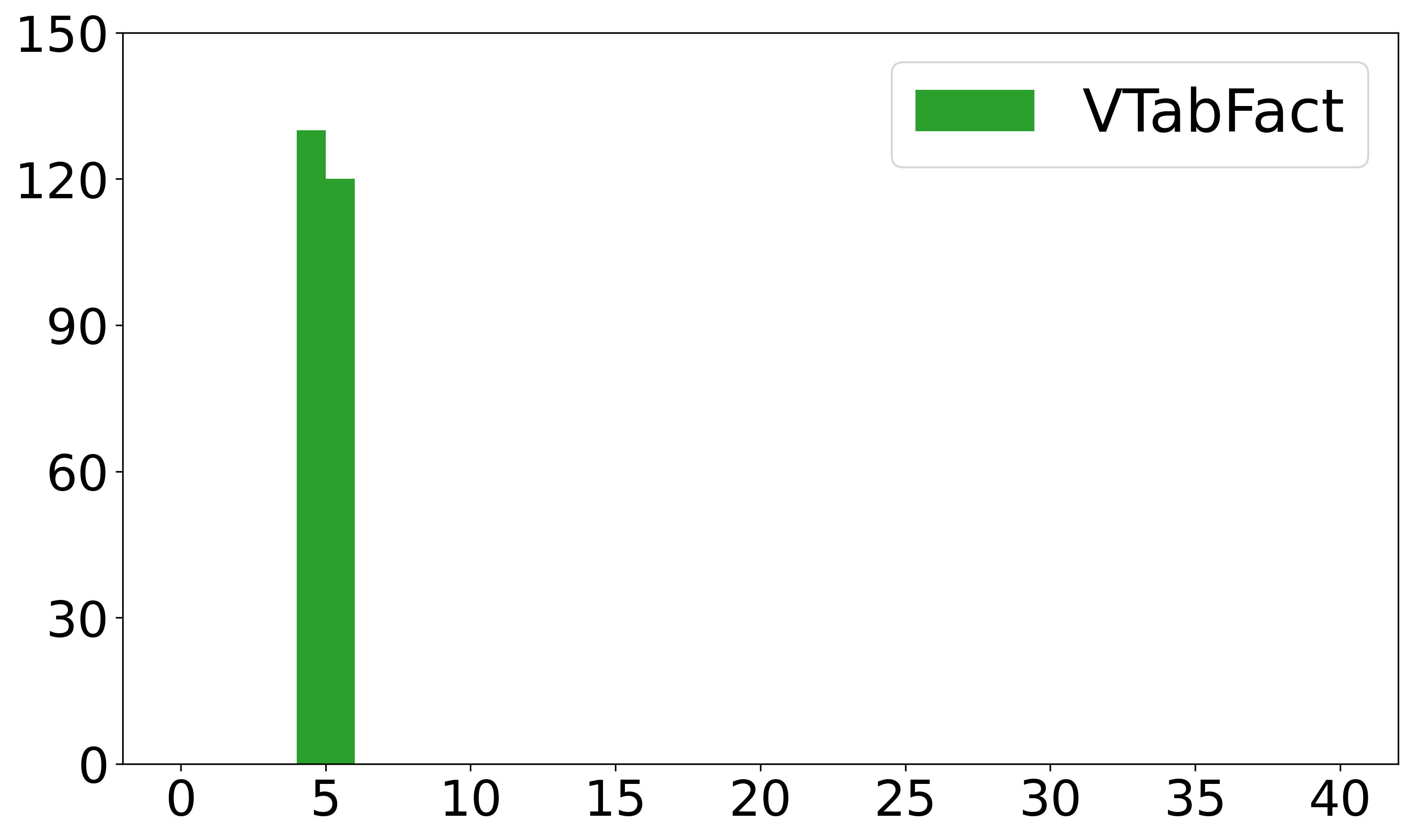} 
        \includegraphics[width=0.24\textwidth]{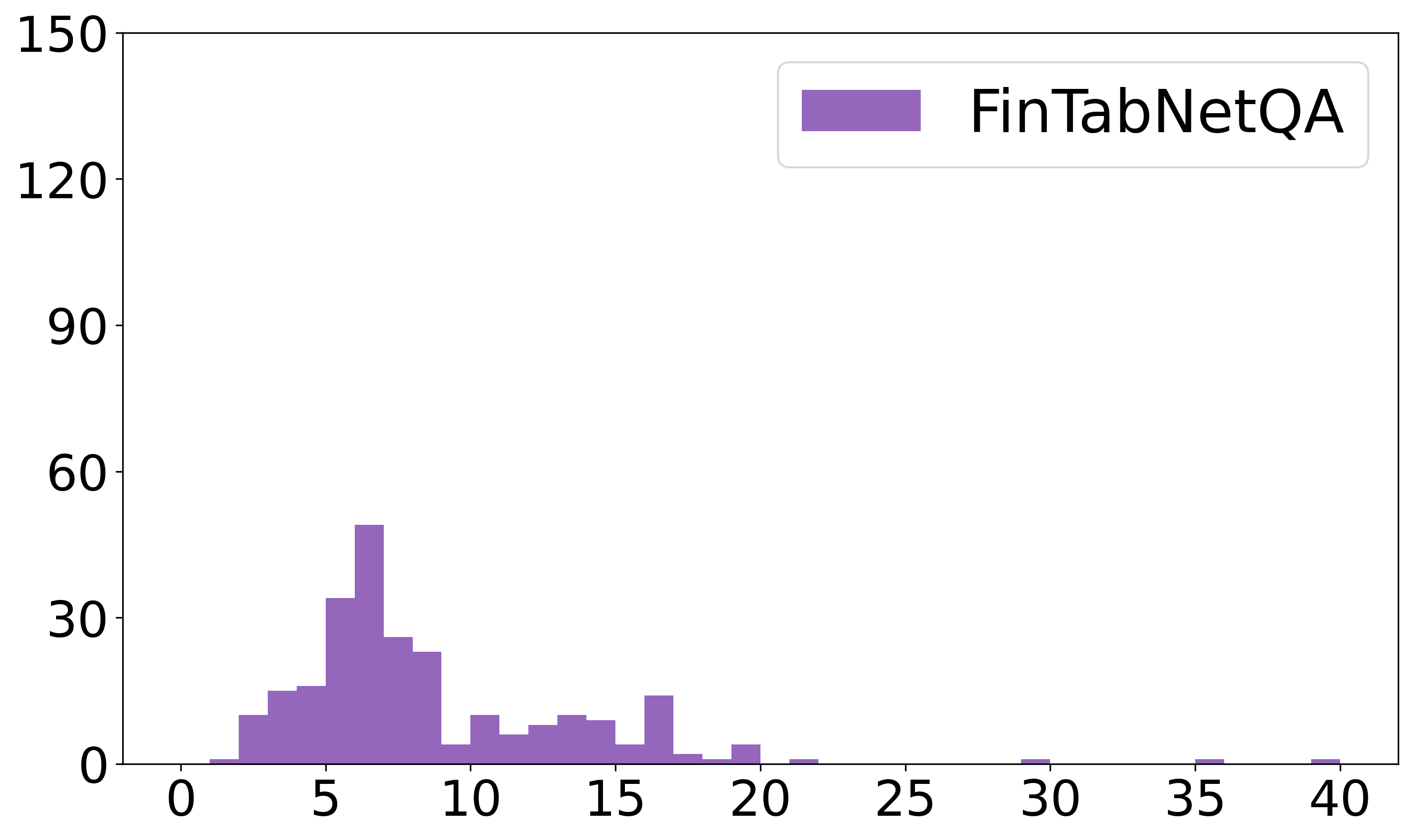}      
        \caption{Answer Length}
        \label{fig:eda_answer}
    \end{subfigure}
    \begin{subfigure}[b]{1.0\textwidth}
        \centering
        \includegraphics[width=0.24\textwidth]{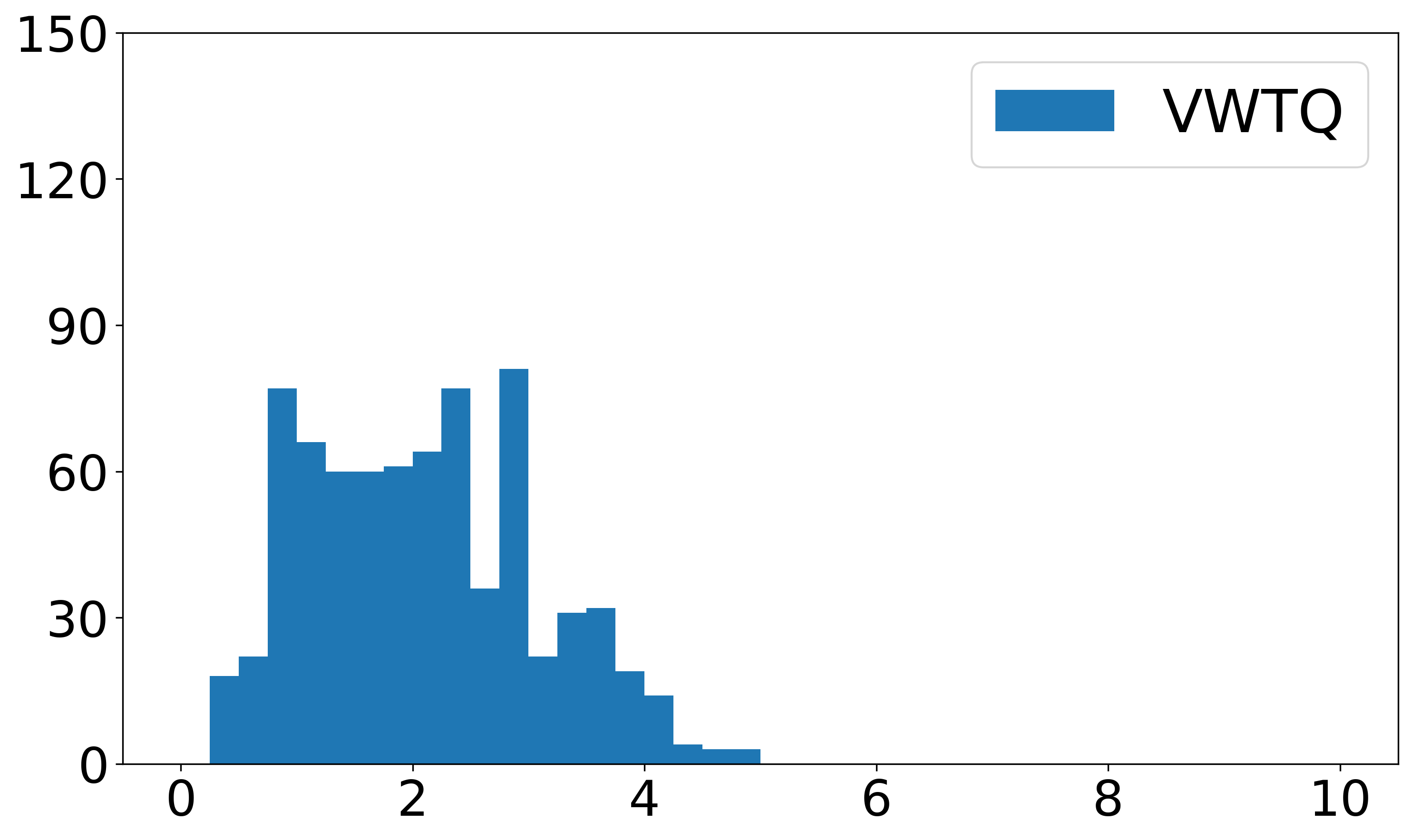}
        \includegraphics[width=0.24\textwidth]{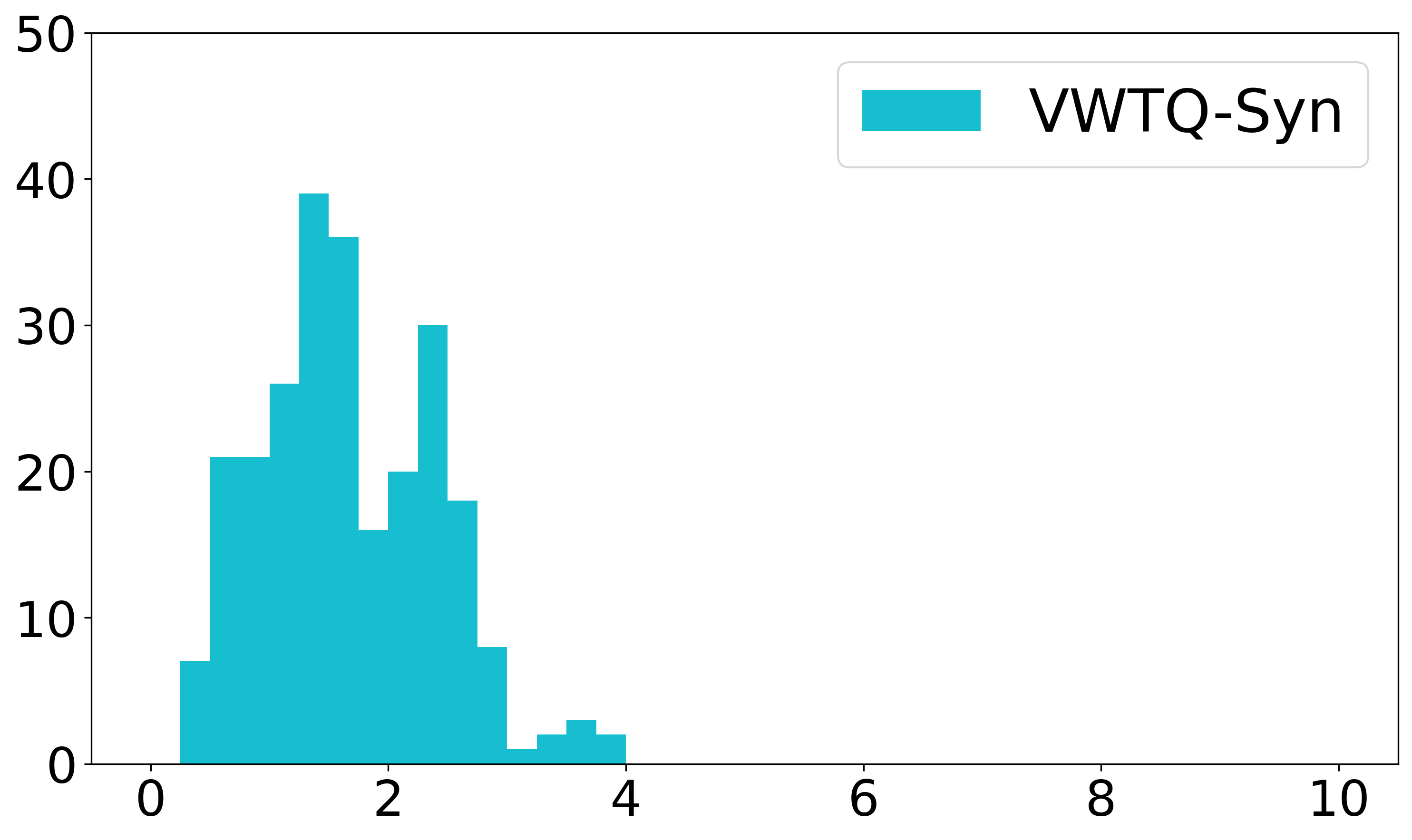} 
        \includegraphics[width=0.24\textwidth]{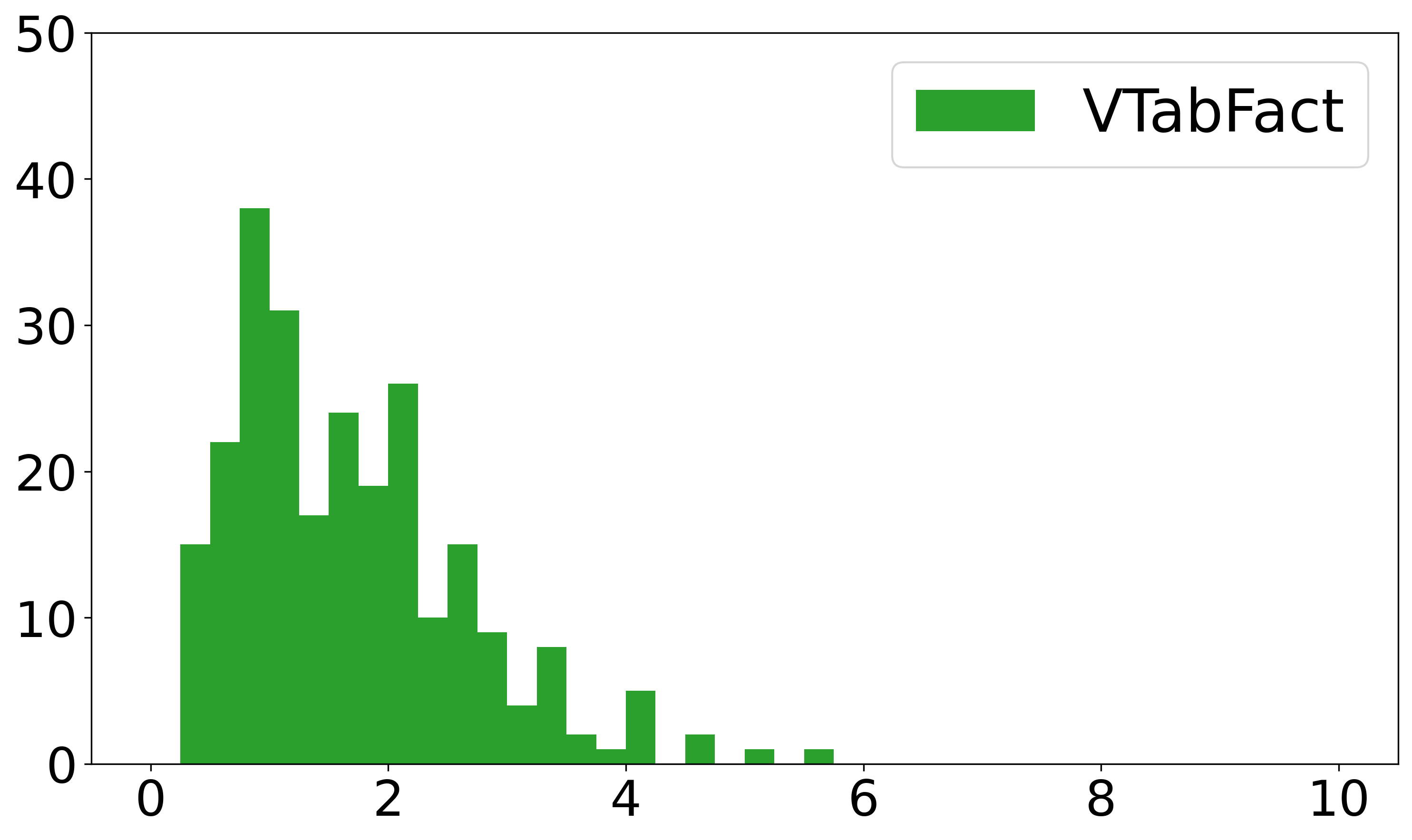} 
        \includegraphics[width=0.24\textwidth]{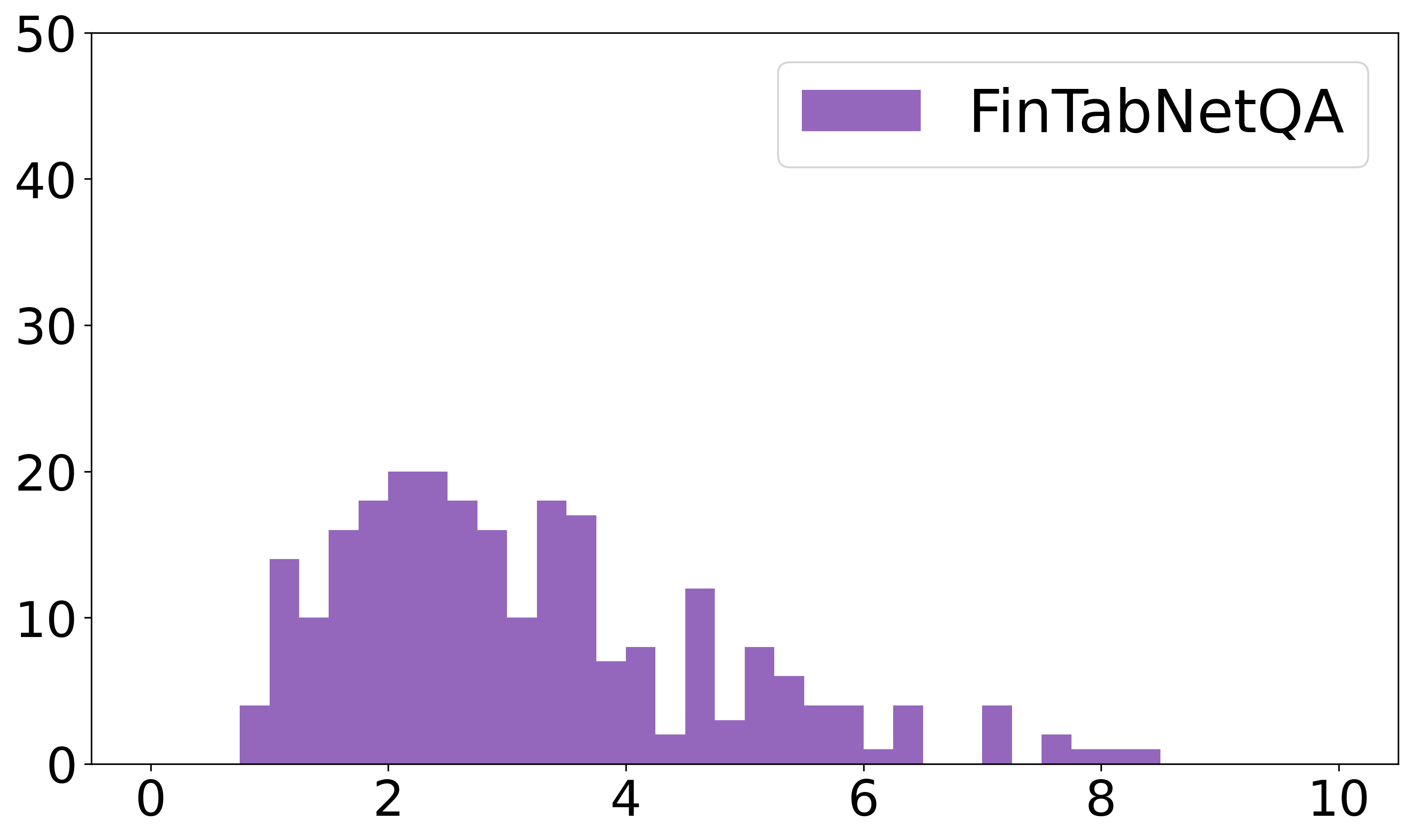}      
        \caption{Aspect Ratio}
        \label{fig:eda_aspect}
    \end{subfigure}
    \begin{subfigure}[b]{1.0\textwidth}
        \centering
        \includegraphics[width=0.24\textwidth]{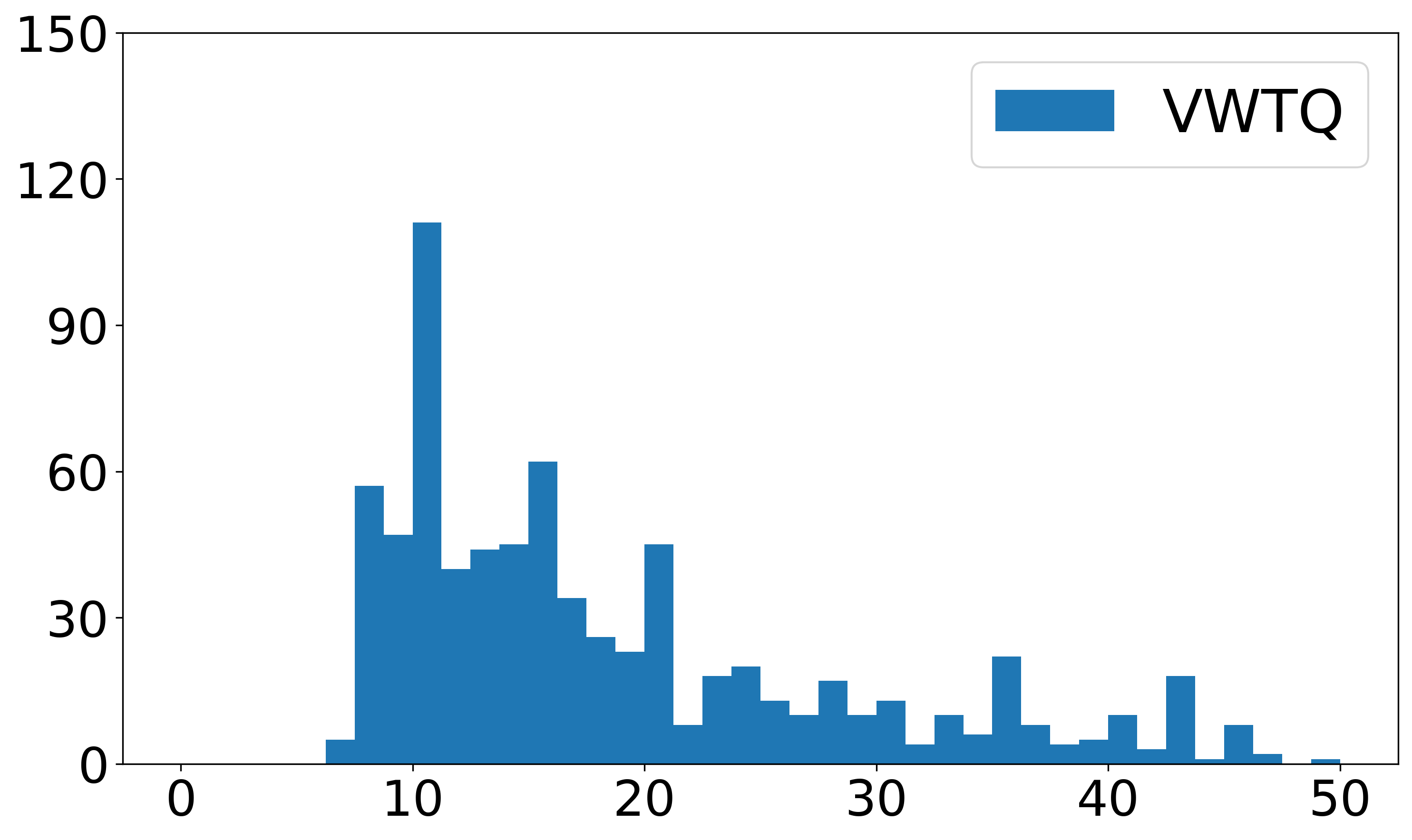}
        \includegraphics[width=0.24\textwidth]{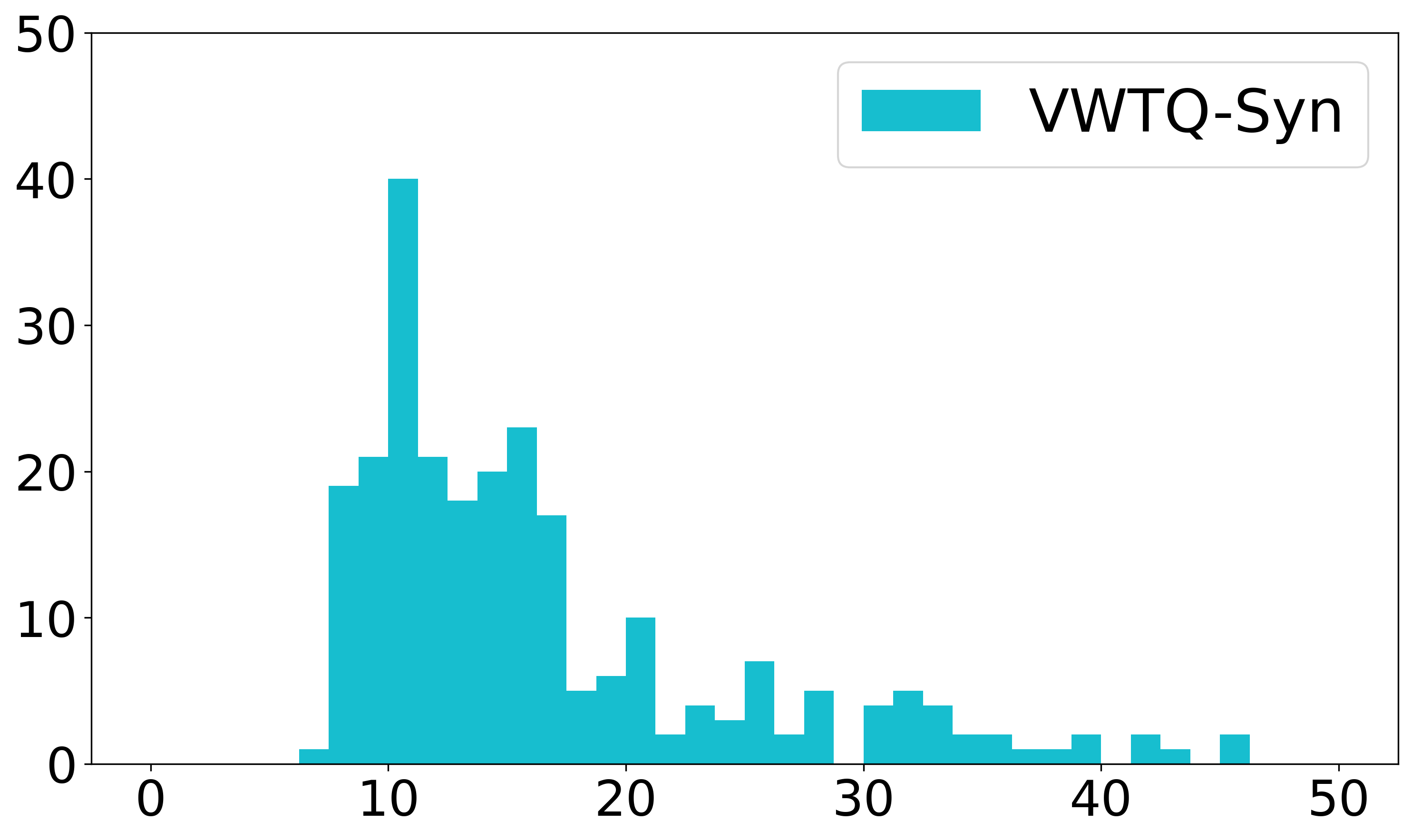} 
        \includegraphics[width=0.24\textwidth]{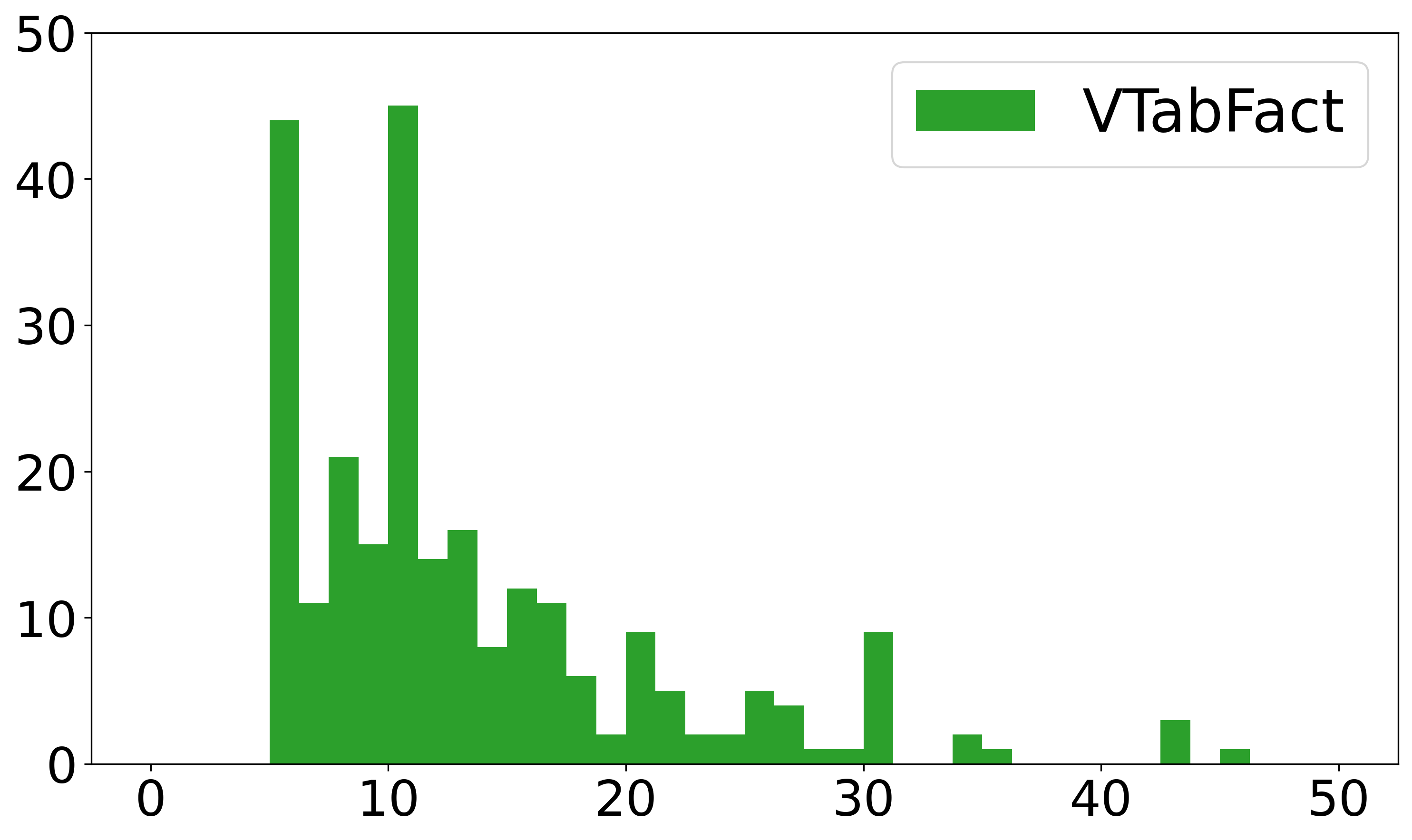} 
        \includegraphics[width=0.24\textwidth]{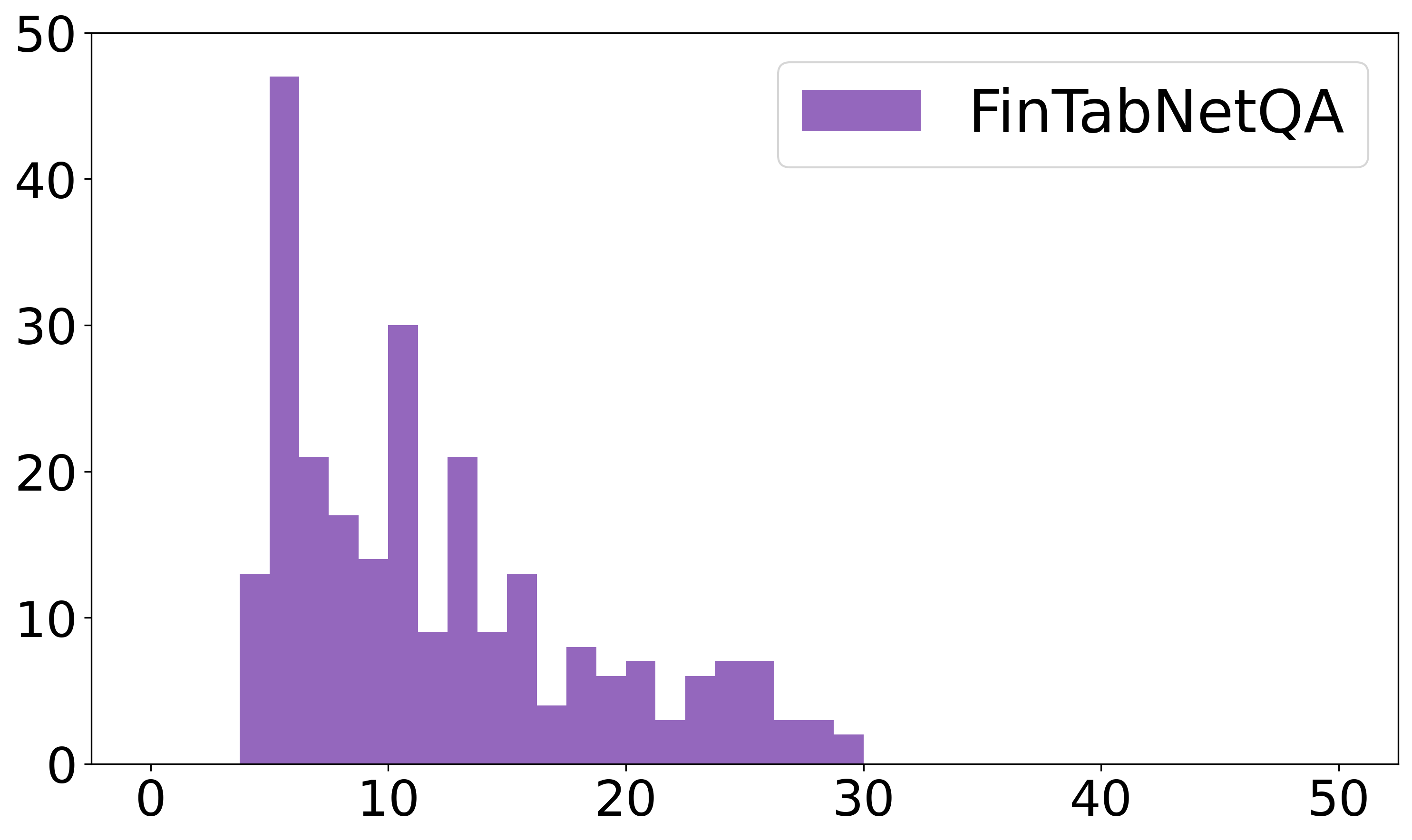}      
        \caption{Number of Rows}
        \label{fig:eda_numrow}
    \end{subfigure}
    \begin{subfigure}[b]{1.0\textwidth}
        \centering
        \includegraphics[width=0.24\textwidth]{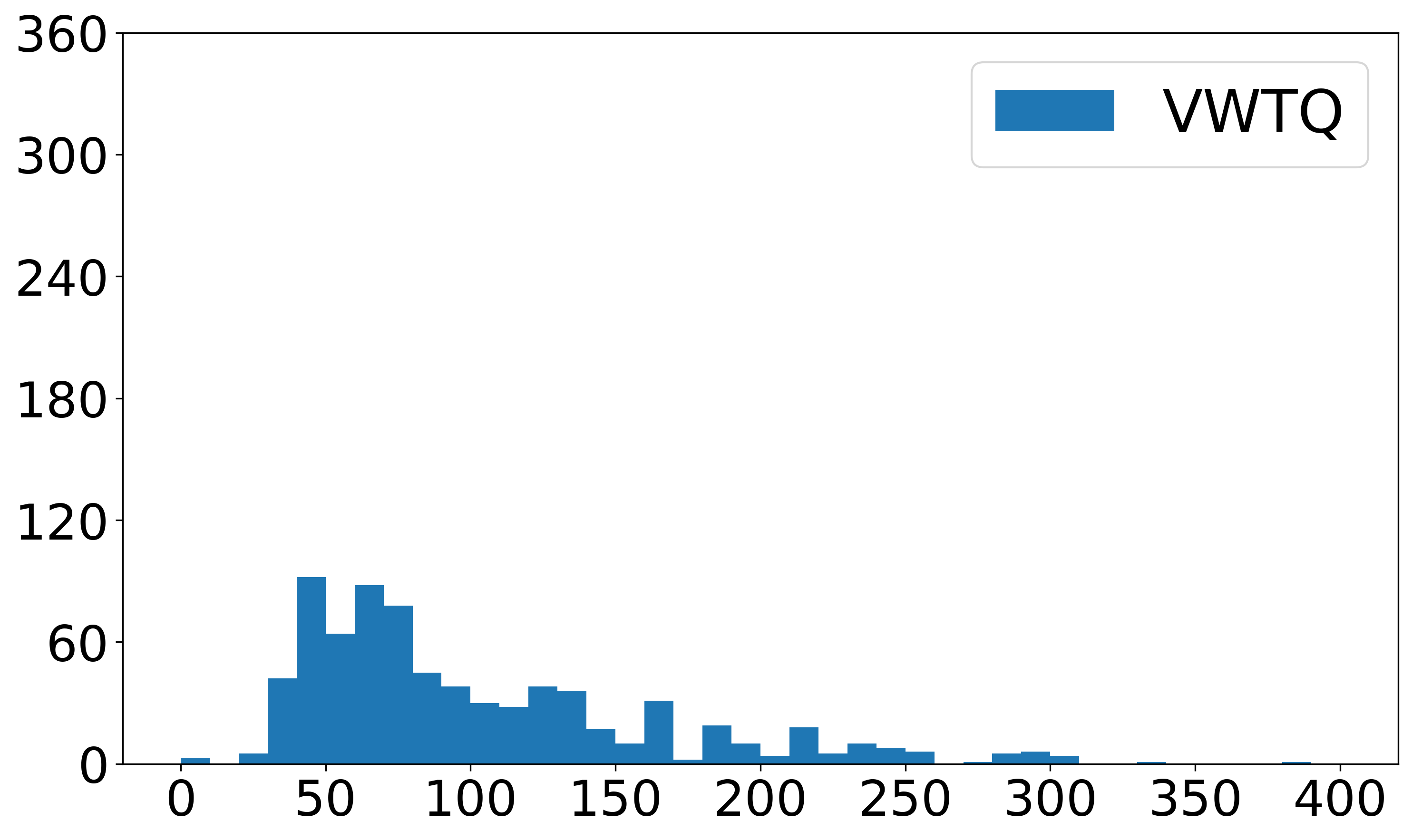}
        \includegraphics[width=0.24\textwidth]{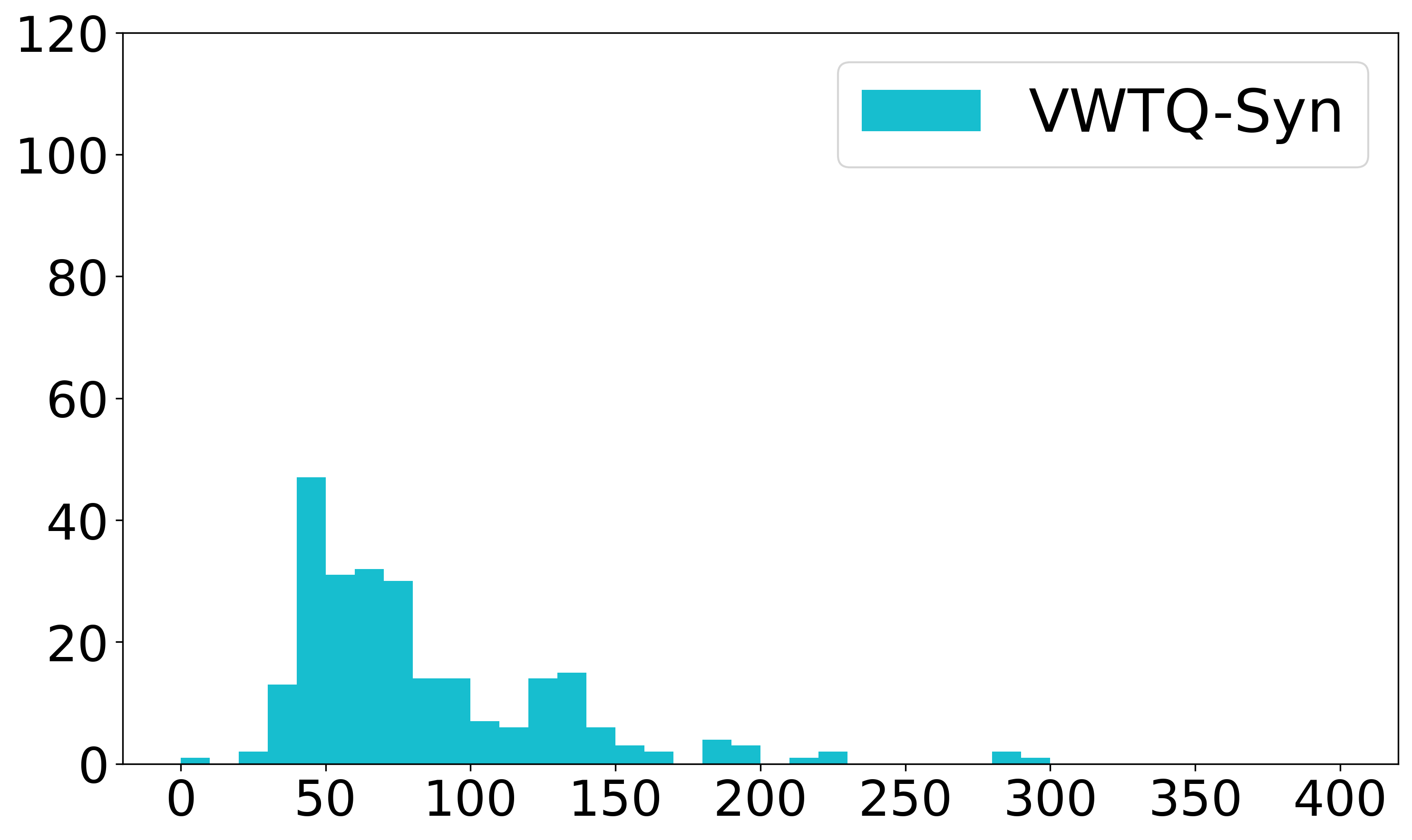} 
        \includegraphics[width=0.24\textwidth]{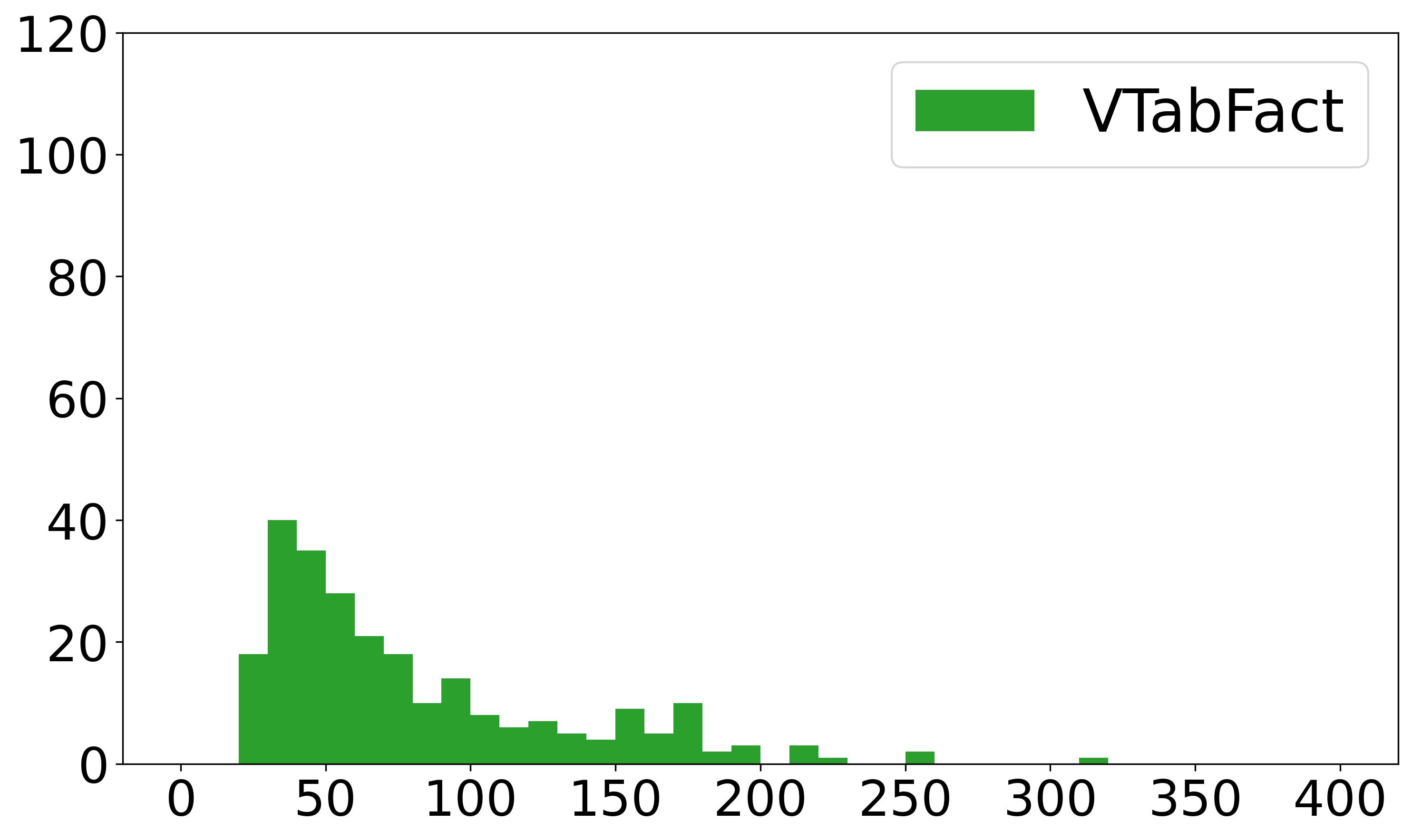} 
        \includegraphics[width=0.24\textwidth]{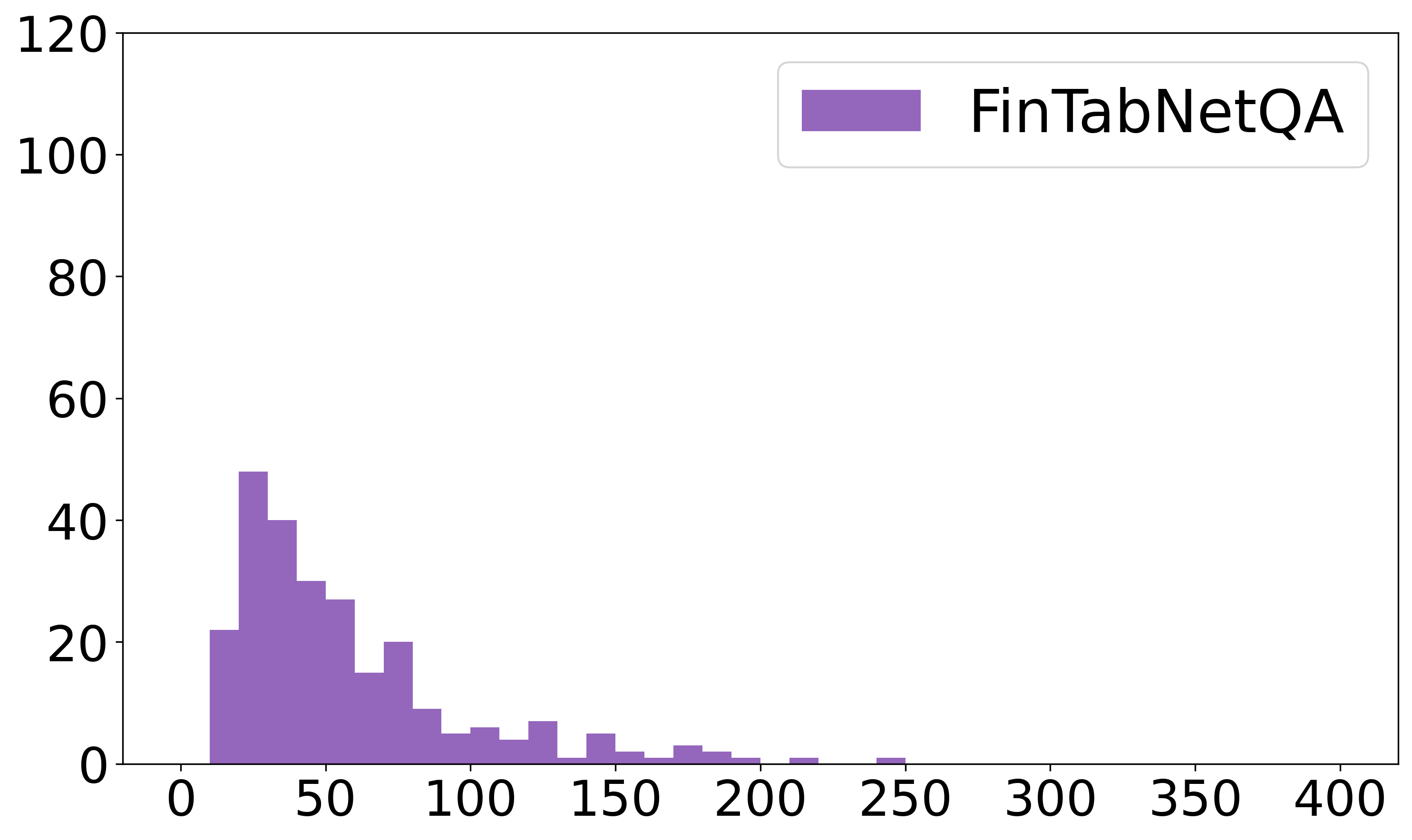}      
        \caption{Number of Cells}
        \label{fig:eda_numcell}
    \end{subfigure}       
    \begin{subfigure}[b]{1.0\textwidth}
        \centering
        \includegraphics[width=0.24\textwidth]{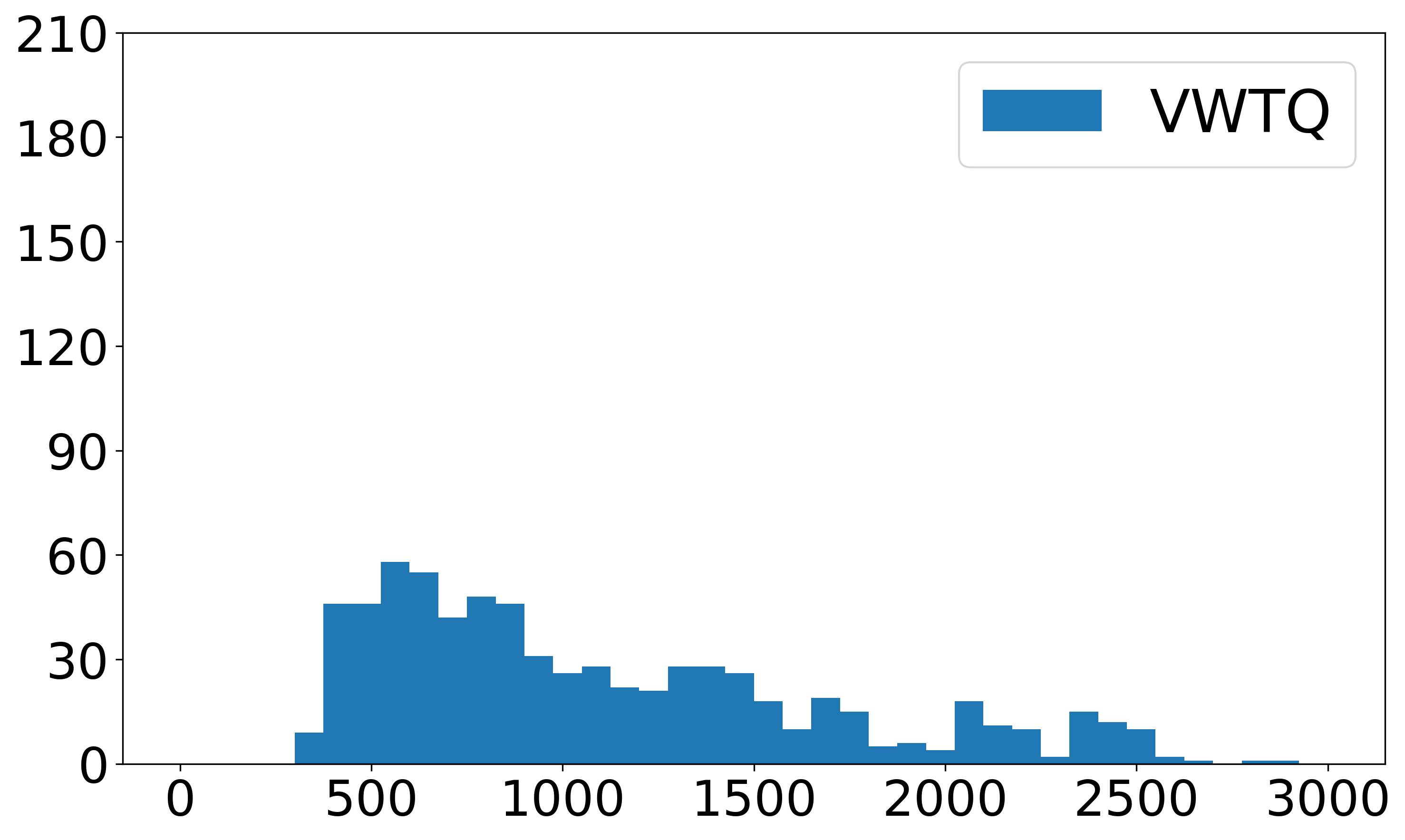}
        \includegraphics[width=0.24\textwidth]{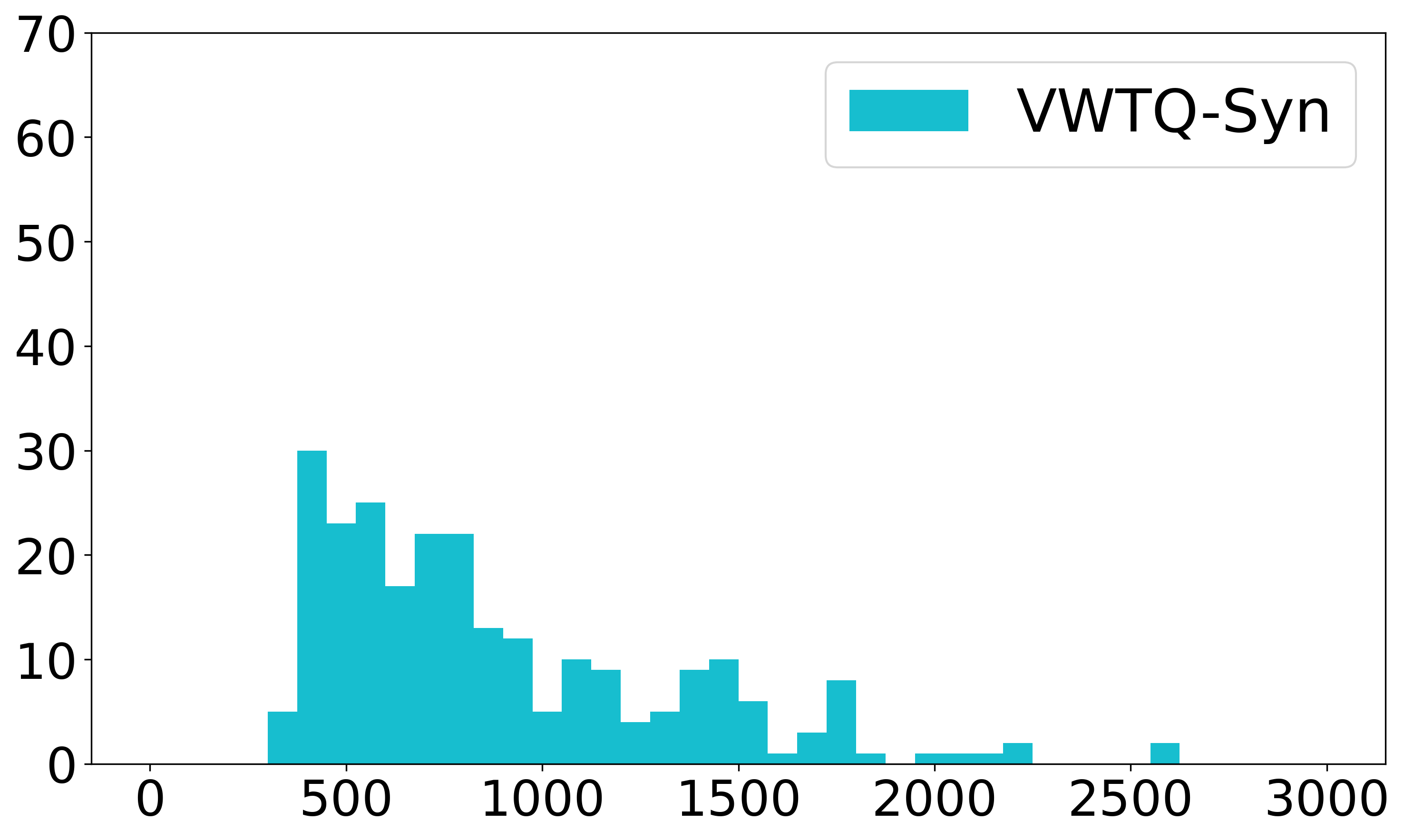} 
        \includegraphics[width=0.24\textwidth]{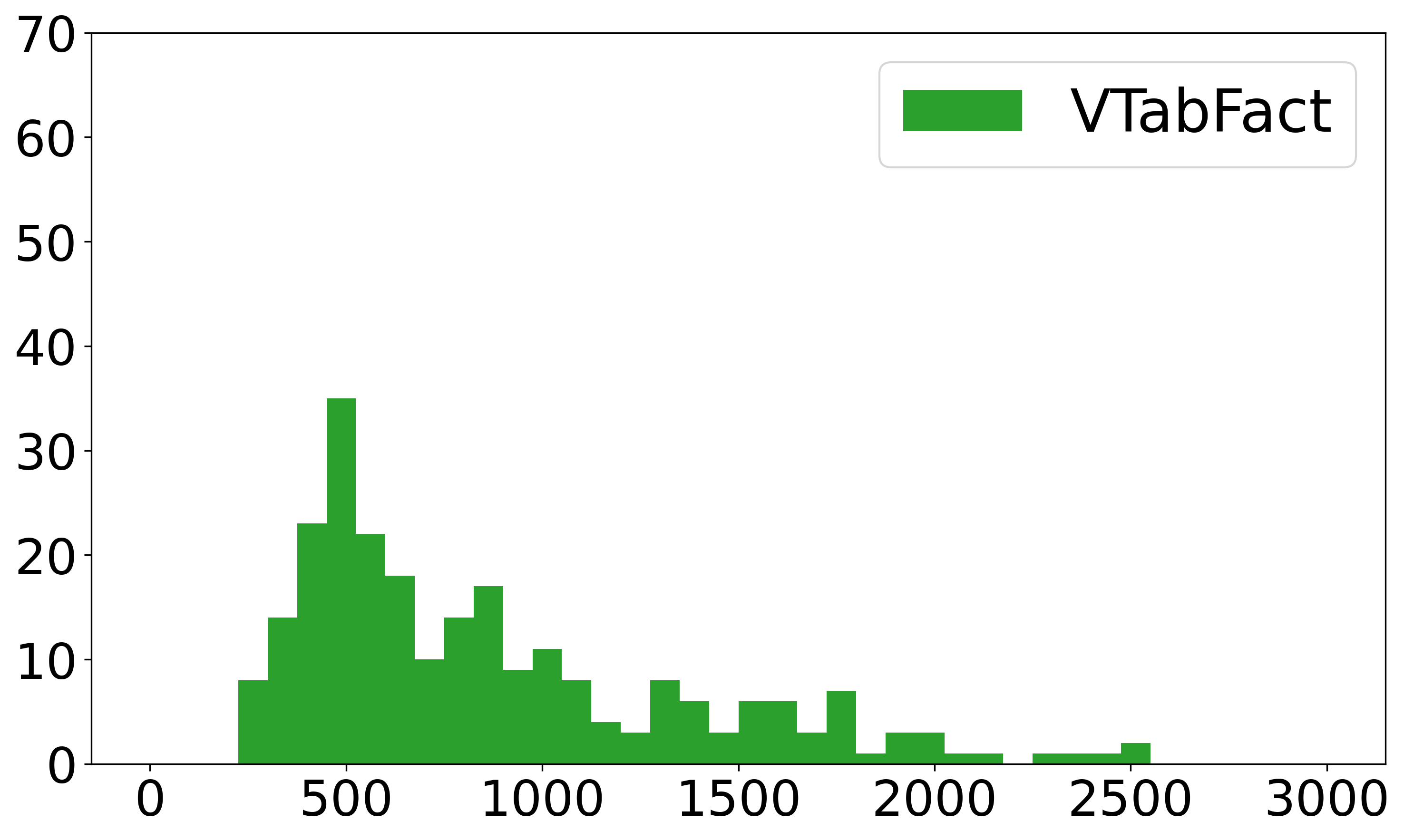} 
        \includegraphics[width=0.24\textwidth]{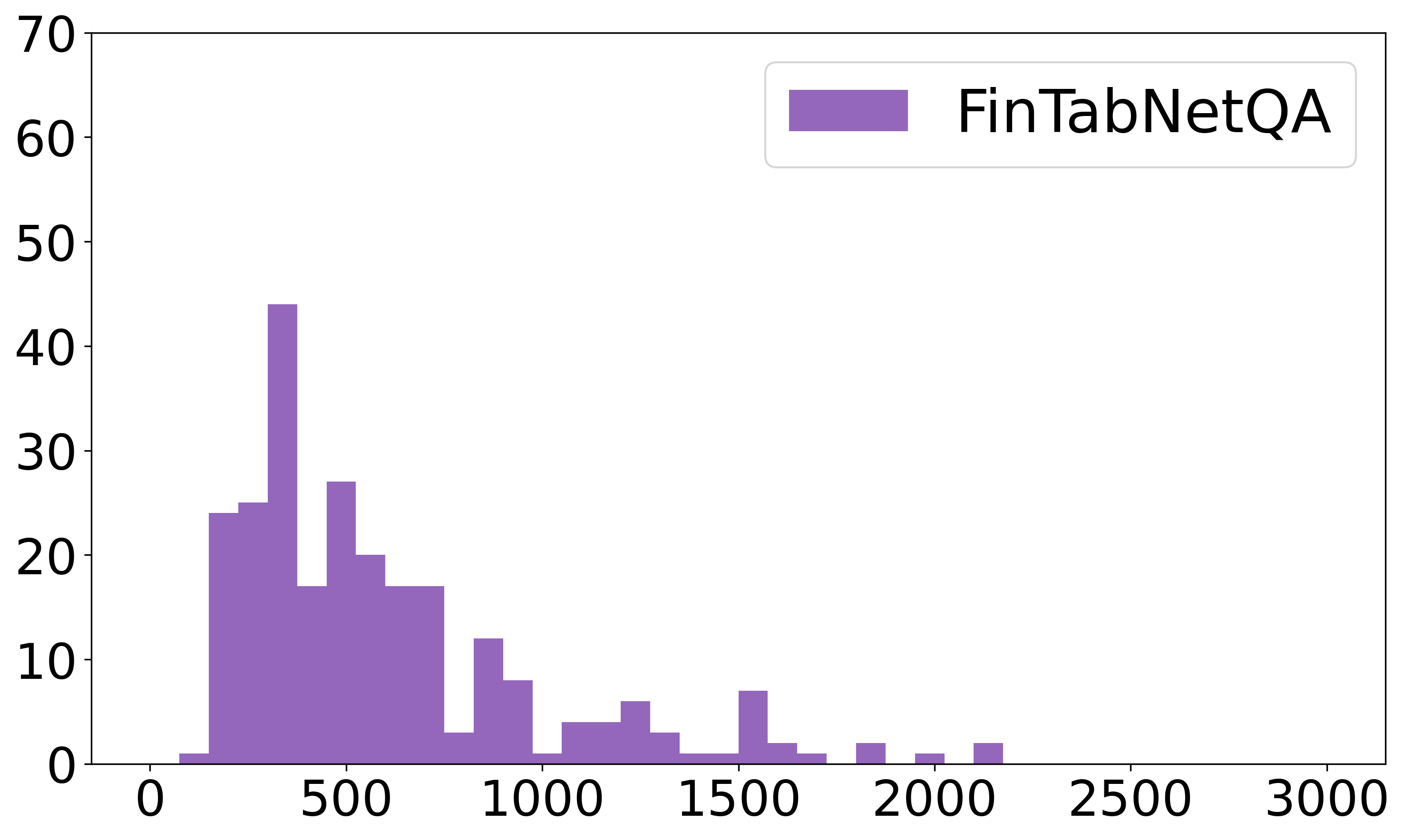}      
        \caption{Number of Text Tokens}
        \label{fig:eda_numtoken}
    \end{subfigure}    
    \begin{subfigure}[b]{1.0\textwidth}
        \centering
        \includegraphics[width=0.24\textwidth]{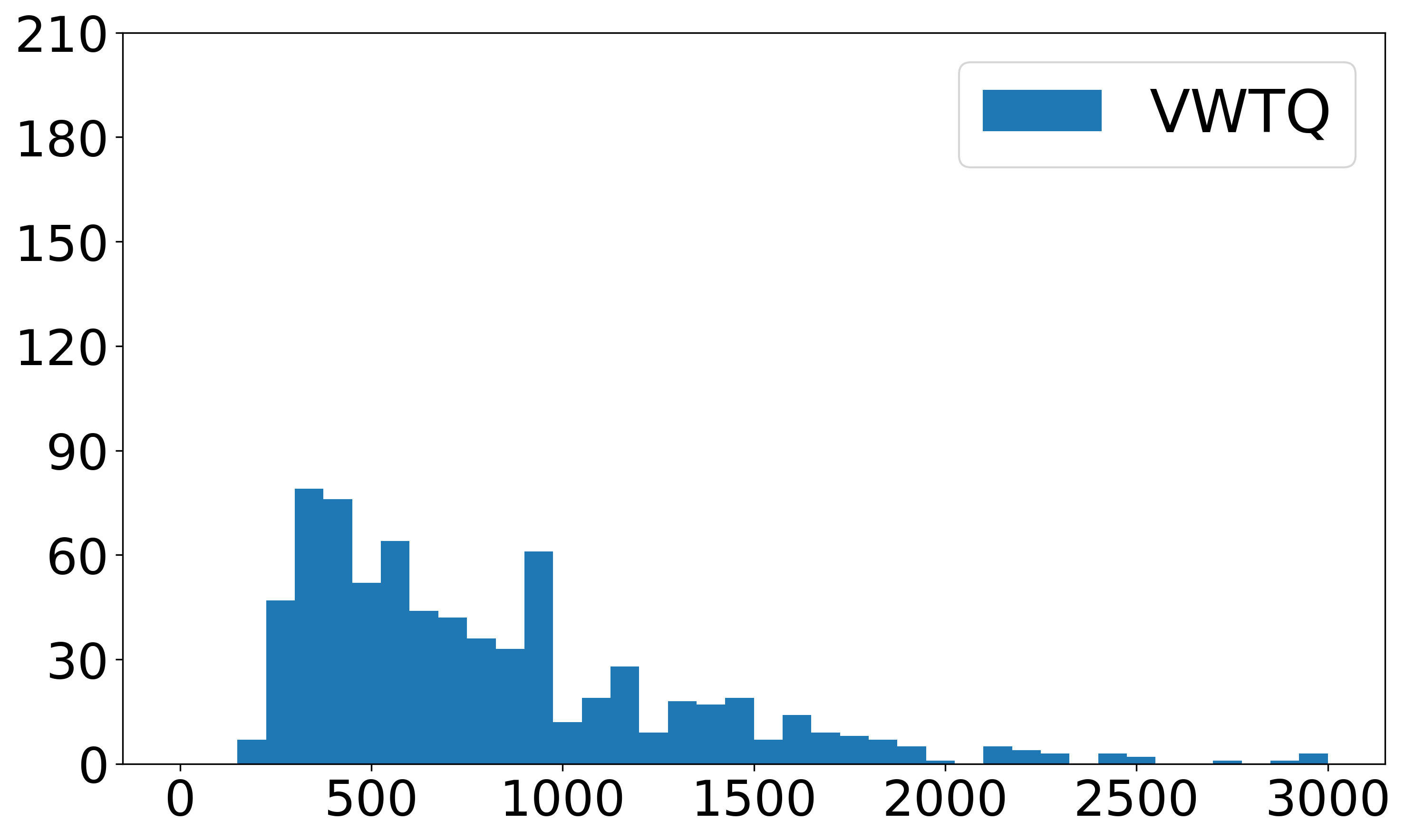}
        \includegraphics[width=0.24\textwidth]{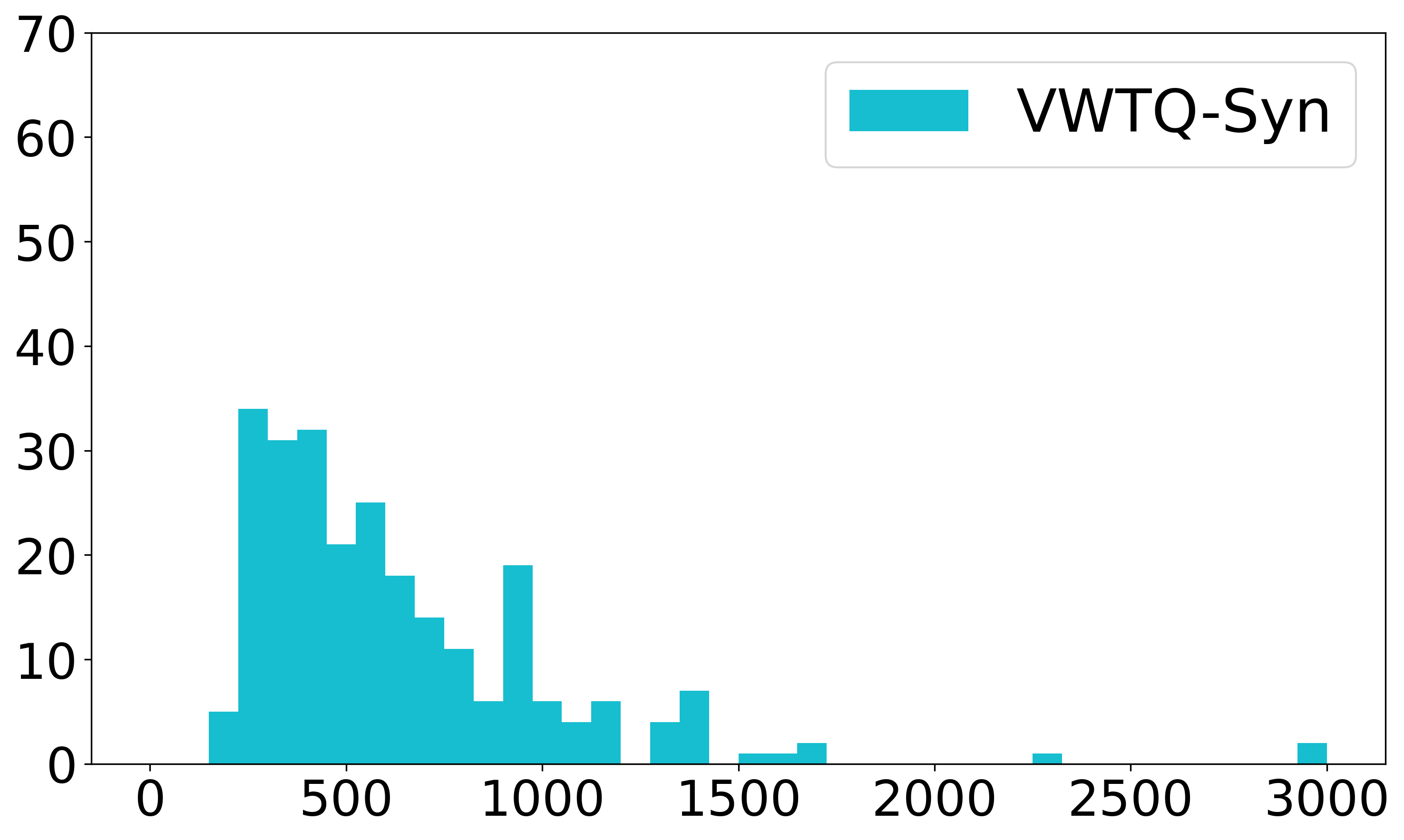} 
        \includegraphics[width=0.24\textwidth]{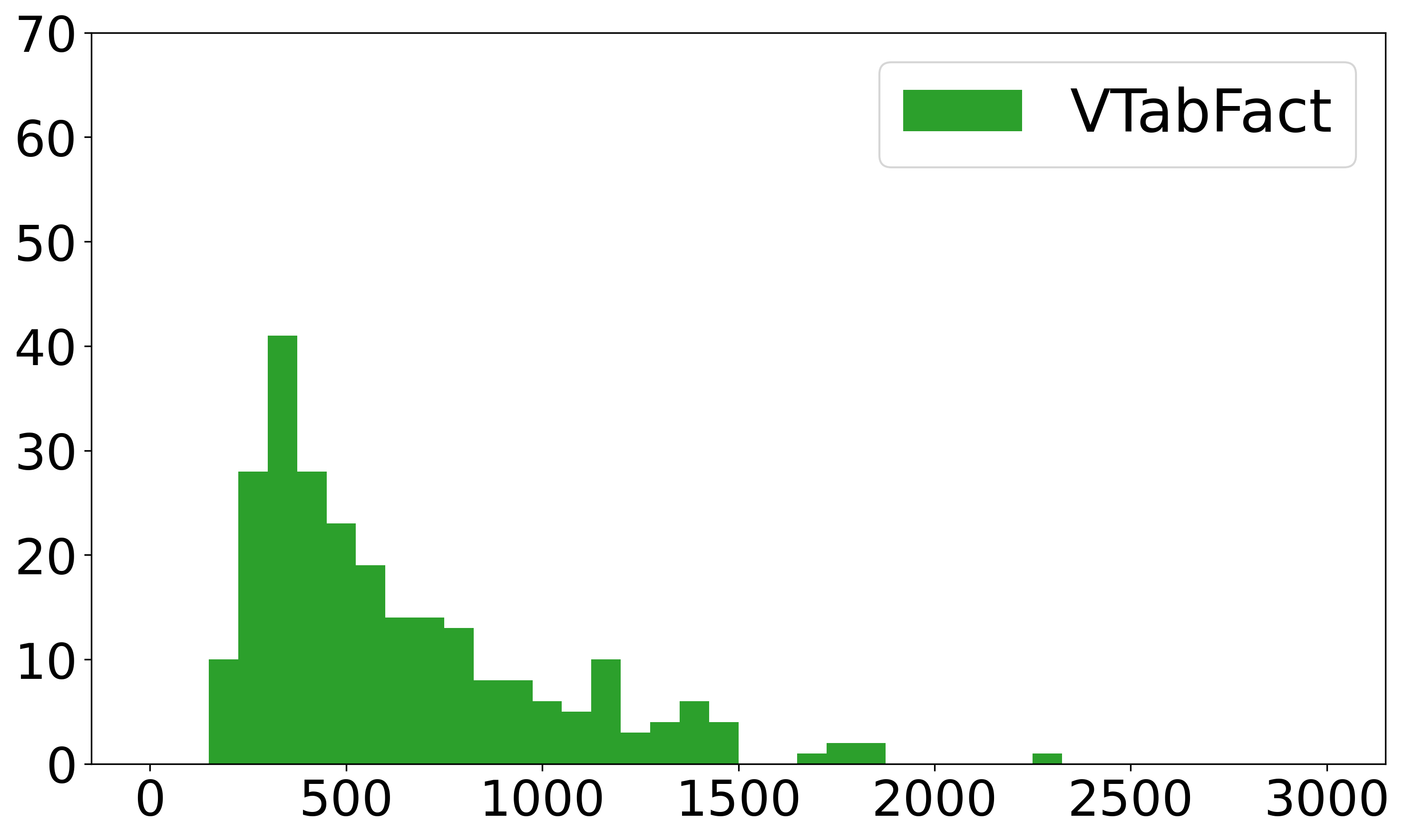} 
        \includegraphics[width=0.24\textwidth]{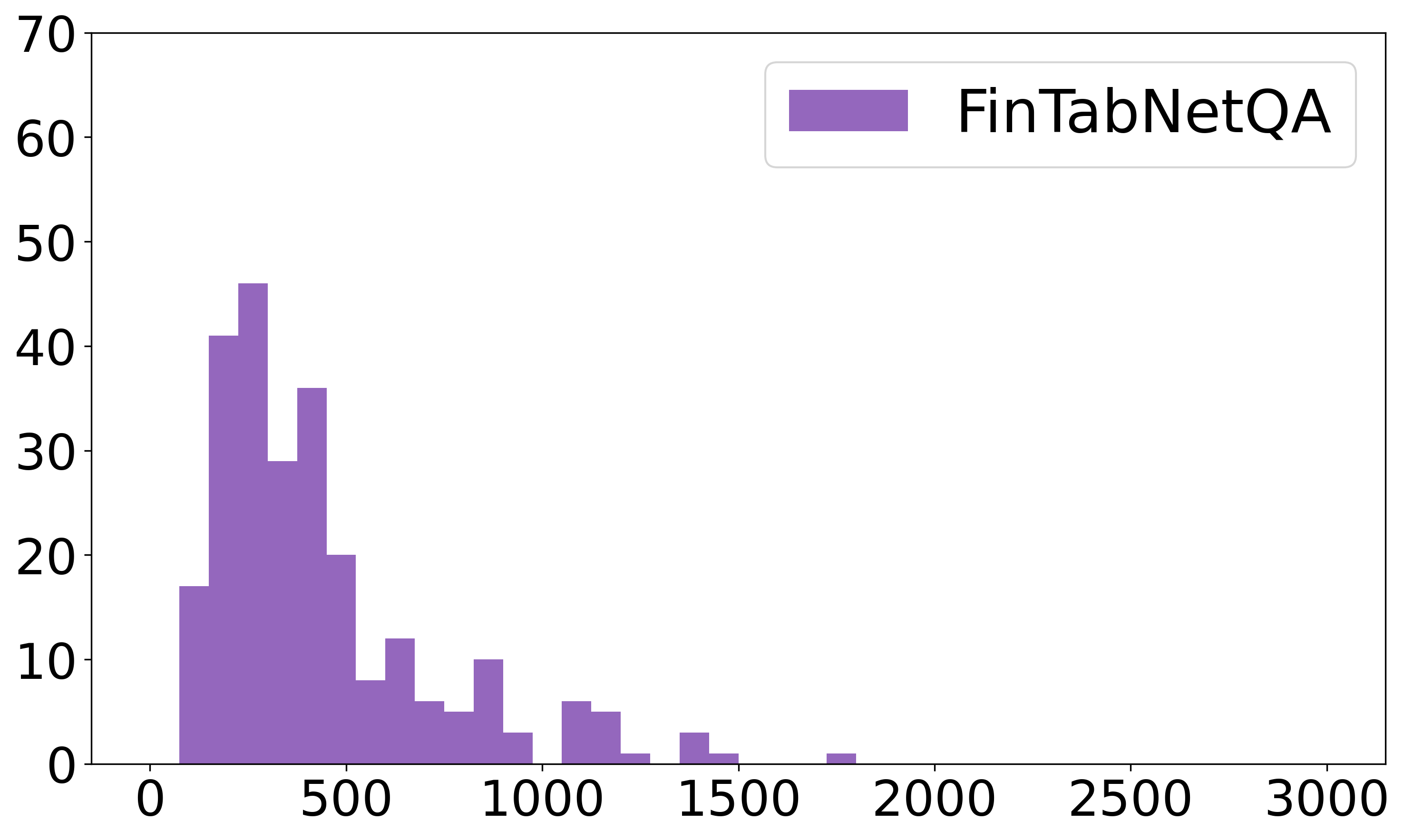}      
        \caption{Number of Text Tokens with Special Tokens}
        \label{fig:eda_numoken_special}
    \end{subfigure}
\caption{The distribution is analyzed with respect to each feature. For the quantification of text tokens, we utilize the Viucuna-7B~\cite{vicuna2023} tokenizer.} \label{fig:eda}
\end{figure}
\subsection{Data Statistics}
Table~\ref{table:data_abstract} provides data statistics, comprising a total of 894 images and 1500 QA pairs for evaluation. VWTQ includes 750 QA pairs gathered from purely authentic data. An equal quantity of QA pairs is amassed from partial real data, originating from VWTQ-syn, VTabFact-syn, and FintabNetQA. The QA pairs of VWTQ-syn are sampled from VWTQ. 

The distribution of each dataset is examined and visualized for analytical purposes in Fig.~\ref{fig:eda}. The observed statistics in Fig.~\ref{fig:eda_question} reveal that the length of the questions originating from FintabNetQA is generally longer than the other datasets. This trend is possibly due to its machine-generated characteristics, where GPT-4 tends to construct more elaborate question structures. As shown in Fig.~\ref{fig:eda_answer}, the answer length distribution for VTabFact seems to branch out into two distinctive categories, with ``true'' or ``false'' being its definitive responses. Frequent instances of elongated answers in FintabNetQA primarily occur due to the common inclusion of units. 
As shown in Fig.~\ref{fig:eda_aspect},~\ref{fig:eda_numrow},  and~\ref{fig:eda_numcell}, a prominent correlation between the number of rows and the aspect ratio can be established. VWTQ is distinctively characterized by the presence of numerous tables with lengthy rows. While comparing the number of rows, FintabNetQA often exhibits a larger aspect ratio. This might be attributed to two possible explanations: 1) the cell height is relatively larger, and 2) the cell content is abundant, leading to an increase in the number of line breaks.

As illustrated in Fig. \ref{fig:eda_numtoken} and \ref{fig:eda_numoken_special}, our analysis extends to examining the token length with the Vicuna-7B tokenizer~\cite{vicuna2023} when tables are encoded in HTML format. We found that the tokenizer does not incorporate HTML tags such as $<td>$, $<tr>$, and $<th>$ as individual tokens. Although incorporating these tags as special tokens slightly increases the vocabulary size, it significantly reduces the number of required input tokens.
Typically, open-sourced MLLMs~\cite{liu2023llava-1.5,blip2} integrate vision queries and text queries by concatenating them before feeding them into the LLMs. 
Consequently, comparing the length of text tokens with the length of vision tokens becomes feasible when tables are represented in both image and text formats.
As shown in Table~\ref{table:compared_models}, the length of vision tokens varies widely, ranging from 32 to 1445. It is observed that the efficiency of image-formatted tables significantly decreases compared to those text-formatted with special tokens when the length of a vision query exceeds 1,000 tokens.

\section{Experiments}


\subsection{Experimental Setup}
\paragraph{\textbf{Evaluation Protocol.}}
In the inference phase, minor prompt tuning was conducted for each model in order to acquire a suitable answer format for subsequent evaluation. In instances where answer parsing was required, rule-based methods are deployed.  
The chosen metric for evaluation is accuracy, the specifics of which are explained in Section 3. 
When the rule-based parsing fails to acquire a properly formatted answer, we also evaluate its performance using a modified accuracy metric.
This metric specifically assesses whether the answer is contained within the response.
These aforementioned processes will be incorporated into the upcoming project page.
\begin{table}[t]
\begin{center}
\caption{The architecture of open-sourced MLLMs. $\alpha$ denotes an additional number of vision tokens that feed to the cross-attention layer of LLM.} \label{table:compared_models}
\resizebox{0.95\linewidth}{!}{%
\begin{tabular}{lc|cc|ccc} \toprule
Models   & Size         & LLM Branch & Size & Vision Branch & Size & \#Vision-Queries\\ \midrule \midrule
BLIP-2~\cite{blip2} & 12.1B & FlanT5-XXL & 11B & EVA-CLIP-g/14 & 1B & 32 \\
InstructBLIP~\cite{instructblip} & 8.2B & Vicuna-7B & 7B &  EVA-CLIP-g/14 & 1B&  32\\
CogVLM~\cite{wang2023cogvlm} & 17B        & Vicuna-7B & 7B  & EVA-02-CLIP-E/14 & 4.4B & 256 \\
CogVLM-1k~\cite{wang2023cogvlm} & 17B        & Vicuna-7B & 7B  & EVA-02-CLIP-E/14 & 4.4B & 1225 \\
CogVLM-Agent-VQA~\cite{cogagent} & 17B        & Vicuna-7B & 7B  & Mixed & 4.4B & 256+$\alpha$ \\
mPLUG-Owl2~\cite{ye2023mplug} & 8.2B       & LLaMA-7B & 7B  & CLIP ViT-L/14 & 0.3B  & 64 \\
SPHINX-v1~\cite{lin2023sphinx}  & 15.7B        & LLaMA-13B & 13B  & Mixed & 2.7B & 289\\
SPHINX-v1-1k~\cite{lin2023sphinx} & 15.7B        & LLaMA-13B & 13B  & Mixed & 2.7B & 1445 \\
LLaVA-v1.5~\cite{liu2023llava-1.5}  & 13.4B & Vicuna-13B & 13B & CLIP ViT-L/14 & 304M & 576 \\
Qwen-VL(-Chat)~\cite{qwen} & 9.6B & Qwen-7B & 7.7B &OpenCLIP ViT-G/14 & 1.9B & 256 \\ \bottomrule  
\end{tabular}}
\end{center}
\end{table}

\paragraph{\textbf{Compared Models.}} 
Comparative analysis is conducted on MLLMs, including commercial models such as Gemini-ProV\footnote{gemini-pro-vision and gemini-pro are employed for MLLM and LLM, respectively.}~\cite{gemini} and GPT-4V\footnote{gpt-4-vision-preview and gpt-4-1106-preview are employed for MLLMs and LLM, respectively. For gpt-4-vision-preview, we adopt `auto' as a detail option}~\cite{gpt4v}, and several open-source models as outlined in Table~\ref{table:compared_models}. 
Since SPHINX-MoE and SPHINX-v2 have not been published, we exploited huggingface models\footnote{https://huggingface.co/Alpha-VLLM/LLaMA2-Accessory/tree/main/finetune/mm/SPHINX}.
To examine the capabilities of their underlying LLMs on TableQA, Vicuna-7B-v1.5~\cite{vicuna2023}, Vicuna-13B-v1.5~\cite{vicuna2023}, Gemini-Pro~\cite{gemini}, GPT-3.5, and GPT-4~\cite{gpt4} are evaluated by feeding them HTML-encoded tables as input.
We also employ two-stage inference methods. We extract the HTML of tables using MLLMs and then conduct the QA task with LLMs where these methods are denoted as GPT-4V $\rightarrow$ GPT-4 and Gemini-ProV $\rightarrow$ Gemini-Pro. We expect this to reveal the correlation between textual and visual modalities.

\subsection{Experimental Results}
We present the comprehensive comparisons of multi-modal inputs in Table~\ref{table:overall_peformance}. The average score is achieved from the sample average. 
\setlength{\tabcolsep}{4pt}
\renewcommand{\arraystretch}{1.3} 
\begin{table}[t!]
    \caption{Accuracy scores on TableVQA-Bench. Scores of both text and vision modalities are reported. 
    The notation `-1k' indicates that the number of vision queries is approximately 1k.   
    CogAgent-VQA* denotes the scores evaluated by the modified accuracy metric.      
    The highest scores in each section are represented in  \textbf{bold}.} \label{table:overall_peformance}
    \begin{center}
        \resizebox{1.0\linewidth}{!}{%
            \begin{tabular}{l|l|cccc|c}  \toprule          
            \textbf{Input Modality}       & \textbf{Model}     &  \textbf{VWTQ} & \textbf{VWTQ-Syn} & \textbf{VTabFact}  & \textbf{FinTabNetQA} & \textbf{Avg.} \\ \midrule 
            \rowcolor{maroon!40}
            \multicolumn{7}{c}{\textit{Multi-modal Large Language Models (MLLMs)}} \\ \midrule 
            \multirow{16}{*}{Vision}         & GPT-4V~\cite{gpt4v}        & \textbf{42.5} & \textbf{52.0} & \textbf{68.0}  & \textbf{79.6} & \textbf{54.5} \\
                                             & Gemini-ProV~\cite{gemini}   & 26.7 & 33.2 & 55.6  & 60.8  & 38.3 \\
                                             & SPHINX-MoE-1k  & 27.2 &	33.6 &	61.6 &	36.0 &	35.5   \\
                                             & SPHINX-v2-1k    & 25.3	 &28.0	& 66.8 &	31.2 &	33.7   \\
                                             & QWEN-VL-Chat~\cite{qwen}   & 19.0 &	23.2 &	60.4 &	29.6 & 28.4  \\ 
                                             & QWEN-VL~\cite{qwen}       & 17.2 &	21.2 &	52.0 &	34.0 & 26.5   \\                        & SPHINX-MoE &               15.3 & 	16.8 &	58.8 &	2.8 &	20.7 \\
                                             & SPHINX-v1-1k~\cite{lin2023sphinx}     & 13.2 &	17.2 &	 58.0 &	3.2 &	19.7    \\
                                             & mPLUG-Owl2~\cite{ye2023mplug}    & 10.7 & 14.4 & 56.8  & 2.8   & 17.7 \\
                                             & LLaVA-1.5~\cite{liu2023llava-1.5}     & 12.4 & 12.4 & 55.6  & 0.8   & 17.7 \\        
                                             & CogVLM-1k~\cite{wang2023cogvlm} & 9.7 &	11.6 &	52.0 &	4.8 &	16.3 \\
                                             & SPHINX-v1~\cite{lin2023sphinx}  &7.1 &	9.6 &	55.2 &	1.2	& 14.5 \\
                                             & CogAgent-VQA~\cite{cogagent}        & 0.3 &	0.8 &	58.4 &	22.8 &	13.8    \\        
                                             & InstructBLIP~\cite{instructblip}  & 5.9 &	6.4 &	50.4 &	0.4 & 12.5    \\ 
                                             & BLIP-2~\cite{blip2}  & 5.2 &	5.6 & 51.6 & 0.4 & 12.2 \\
                                             & CogVLM~\cite{wang2023cogvlm} &0.8	&0.8	& 40.8 &	1.2 &	7.5 \\                   
                                             & CogAgent-VQA*~\cite{cogagent} & 37.2 &	41.2 &	58.4 &	22.8 &	39.0 \\ \midrule
            \rowcolor{maroon!40}
            \multicolumn{7}{c}{\textit{Table Structure Reconstruction} + \textit{Large Language Models (LLMs)}} \\ \midrule
            \multirow{2}{*}{Vision}
            & GPT-4V~\cite{gpt4v} $\rightarrow$ GPT-4~\cite{gpt4} & \textbf{45.2} &	\textbf{55.6} &	\textbf{78.0} &	\textbf{95.2} &	\textbf{60.7}  \\
            & Gemini-ProV $\rightarrow$ Gemini-Pro~\cite{gemini} & 34.8 &	40.4 &	71.0 &	75.6 &	48.6  \\  \midrule  
            \rowcolor{maroon!40}
            \multicolumn{7}{c}{\textit{Large Language Models (LLMs)}} \\ \midrule
            \multirow{5}{*}{Text}            & GPT-4~\cite{gpt4}          & \textbf{68.1} & \textbf{69.6} & \textbf{80.0} & \textbf{98.8}  &\textbf{75.5} \\
                                             & Gemini-Pro~\cite{gemini}    & 56.4 & 61.2 & 69.6  & 96.4  & 66.1   \\   
                                             & GPT-3.5        & 50.5	& 54.4 &	68.0 &	93.2 &	61.2\\                                     
                                             & Vicuna-13B~\cite{vicuna2023} & 32.8	& 39.2 &	57.6 &	84.8 &	46.7 \\
                                             & Vicuna-7B~\cite{vicuna2023} & 21.5 &	34.4 &	54.0 &	68.8 &	37.0 \\ \bottomrule
            
            \end{tabular}
            }
    \end{center}
\end{table}

\paragraph{\textbf{Comparisons between MLLMs.}}
Among MLLMs, commercial models outperform open-source alternatives. To elaborate further, the high performance of GPT-4V can be attributed to the use of GPT-4 in creating QA in FintabNetQA. 
However, GPT-4V demonstrates the highest performance across all datasets, not just this specific instance. 
On TableVQA, we also find that the pivotal role is played by the number of vision queries. 
In a specific comparison, SPHINX-MoE-1k, SPHINX-v1-1k, and CogVLM-1k surpass SPHINX-MoE, SHPHINX, and CogVLM, respectively.  
These findings, along with observations from Fig.~\ref{fig:eda_numoken_special}, indicate that vision input generally requires a higher number of queries than text input to achieve promising performance.
Notably, despite LLaVA-1.5 has not been trained on OCR-abundant documents, it exhibits competitive performance to models that included such documents in their training sets.


\paragraph{\textbf{MLLMs vs. LLMs.}} 
From a performance perspective, the text modality outperforms the vision modality as an input source. Specifically, on average, GPT-4 achieves a performance enhancement of 21 \% points more than GPT-4V, while Gemini-pro outperforms Gemini-proV by 27.8 \% points. 
Similarly, open-sourced MLLMs generally have lower performance than their backbone LLMs such as Vicuna-7B and Vicuna-13B.  
Although the spatial information in vision inputs might enable easier comprehension of the instance's location relation, a performance critically dependent on the aspect ratio cannot be overlooked, as seen in Fig.~\ref{fig:input_format_performance}. 
Such findings indicate that in terms of performance, using text inputs still might be advantageous if both vision and text tables are presented. 
Meanwhile, even in non-GPT models such as Gemini-Pro and Vicuna-13B, a high level of performance is obtained on FintabNetQA, suggesting that the inherent complexity of the QA pair in the dataset is relatively low.

\setlength{\tabcolsep}{4pt}
\renewcommand{\arraystretch}{1.3} 
\begin{table}[t]

\begin{center}
\caption{The performance of TSR. TEDs evaluates the scores of both the structure and content of the table. A higher value indicates better performance.} \label{table:teds}
\resizebox{0.7\linewidth}{!}{%
\begin{tabular}{l|cccc|c} \toprule
            & VQWTQ & VWTQ-Syn & VTabFact & FinTabNetQA & Avg.\\ \midrule \midrule
TSR SoTA~\cite{scob} & \textbf{89.7} &	\textbf{84.5} &	\textbf{76.8} &	52.0 &	\textbf{80.4} \\
Gemini-ProV & 72.7 &	78.4 &	73.0 &	65.8 &	72.6 \\
GPT-4V & 64.0 &	76.7 &	72.8 &	\textbf{72.6} &	69.0 \\

\bottomrule  

\end{tabular}}
\end{center}
\end{table}
\begin{figure}[h!]
    \centering      
    \includegraphics[width=0.9\textwidth]{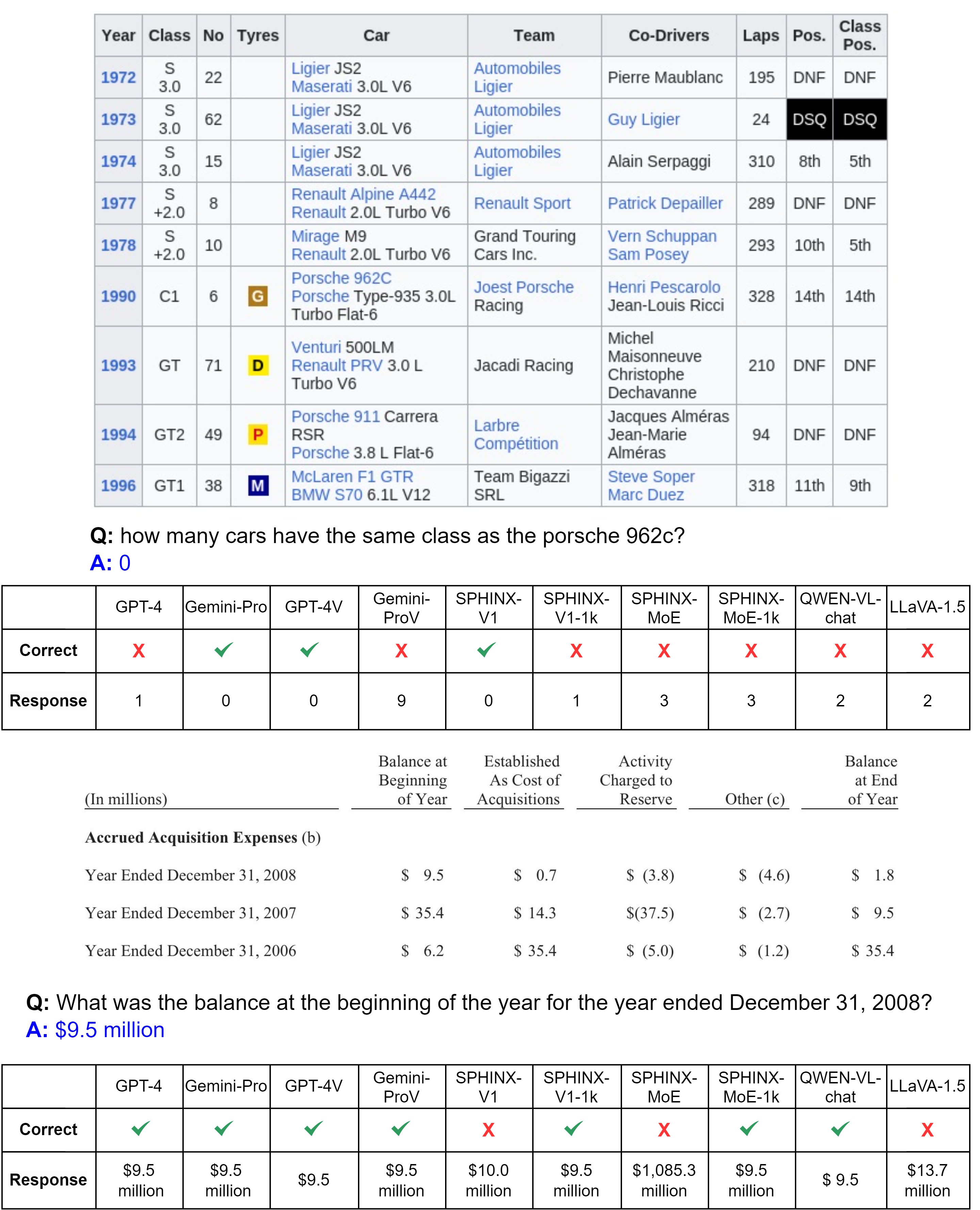}
    \caption{Examples of qualitative evaluation. The examples are sampled from VWTQ (top) and FinTabNetQA (bottom). FinTabNetQA is evaluated with the \textit{relieved-accuracy} where scale units are intentionally excluded at the evaluation.}
    \label{fig:response}  
\end{figure}
\paragraph{\textbf{Two-stage Inference.}} 
Two-stage inference leads to significant performance enhancements within the same vision input on both GPT and Gemini families. Despite such enhancements, it is evident that the performance still falls short compared to when text input is used. 
While it might be feasible to conduct experiments extracting HTML and answers through prompt tuning in the single MLLM, unfortunately, we were unable to obtain results in our desired format. 
Employing TEDs~\cite{zhong2020image} evaluation metric, we compare the MLLMs' performance on TSR  with that of the state-of-the-art (SoTA) model~\cite{scob}. 
For a fair comparison, we utilize the SoTA model trained only on PubTabNet~\cite{pubtabnet}, which can be regarded as a held-out dataset for TableVQA-Bench.
As shown in Table~\ref{table:teds}, the SoTA model usually performs better than MLLMs. 
These findings indicate that MLLMs exhibit limitations in efficiently extracting information from visual tables.


\paragraph{\textbf{Qualitative Evaluation.}} 
We present qualitative results in Fig.~\ref{fig:response}. The incorrect answers are usually derived from words not presented in the table, which may be attributed to the limitations of OCR capability. A longer length of the vision query appears to alleviate these issues, as demonstrated by the correct answers in the second example.

\begin{figure}[t]
\centering       
\includegraphics[width=0.75\textwidth]{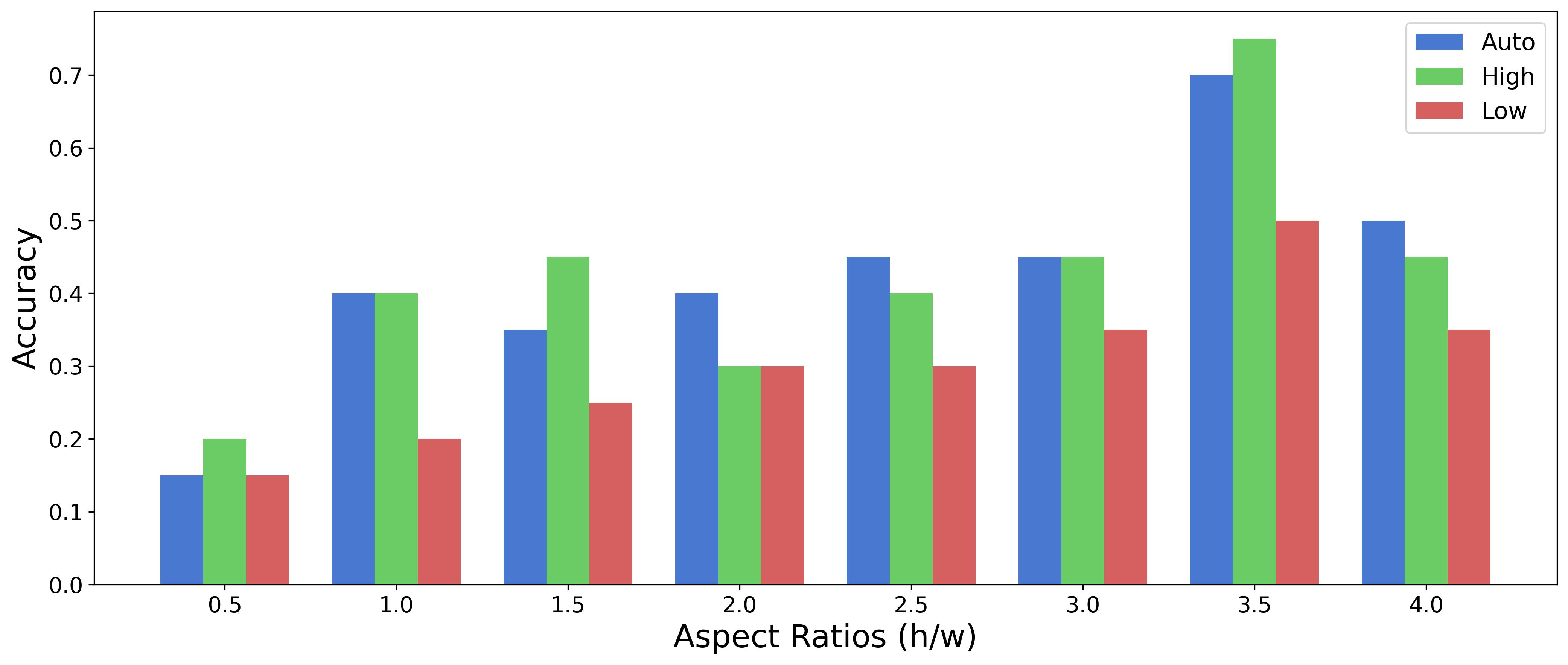}  
\caption{The evaluation is conducted on VWTQ, with 20 instances for each aspect ratio. GPT-4V offers three input image resolution options: `auto', `high', and `low'. The `high' setting requires more computational resources for inference compared to the `low'.}
\label{fig:gpt4v_details}   
\end{figure}
\paragraph{\textbf{GPT-4V Details.}} 
The size of table images can vary significantly depending on their content. In this experiment, we explored the impact on model performance when preserving or not preserving the original size of table images. The GPT-4V offers a `high' option that preserves the input resolution, in contrast to a `low' option that appears to resize the image to a fixed size without preserving the original resolution. Additionally, an `auto' option exists that adaptively determines the resolution based on the input image. For each image ratio, we sampled 20 instances and then measured the performance across these resolution modes. As can be seen in Fig.~\ref{fig:gpt4v_details}, the `low' demonstrated relatively lower performance. 
Hence, maintaining the original resolution constitutes a critical factor for accuracy, which is similarly observed in comparisons among MLLMs.

\section{Conclusion}
In this paper, we present the TableVQA-Bench, a comprehensive benchmark specifically designed for evaluating table visual question-answering capabilities.
To ensure a wide-ranging domain, we have leveraged a multitude of pre-existing table-related tasks, procuring essential elements such as images and question-answer pairs.
Our study includes an extensive evaluation of various models on the TableVQA-Bench. 
Through a comparison among MLLMs, it was observed that GPT-4V outperformed other methods across all evaluated domains. 
Based on observations from the comparison with LLMs and the application of a two-stage inference approach, we believe there is significant potential for further enhancements in MLLMs' performance on visual table understanding tasks.

\paragraph{Acknowledgements}
We greatly appreciate Bado Lee and YoungSang Yoo for their help with the initial project setup.

\bibliographystyle{splncs04}
\bibliography{reference}

\end{document}